\documentclass[10pt,journal,compsoc]{IEEEtran}
\newcommand{\xjqi}[1]{{\color{magenta}{ [xjqi: #1]}}}
\newcommand{\lzz}[1]{{\color{cyan}{\bf\sf [lzz: #1]}}}
\newcommand{\phil}[1]{{\color{red}{[ph: #1]}}}

\newcommand{\ie}{\textit{i.e.}}
\newcommand{\eg}{\textit{e.g.}}

\usepackage[utf8]{inputenc} 
\usepackage[T1]{fontenc}    
\usepackage{hyperref}       
\usepackage{url}            
\usepackage{booktabs}       
\usepackage{amsfonts}       
\usepackage{nicefrac}       
\usepackage{microtype}      
\usepackage{xcolor}         
\usepackage[T1]{fontenc}
\usepackage[linesnumbered,boxed,ruled,commentsnumbered]{algorithm2e}
\usepackage{times}
\usepackage{stfloats}
\usepackage{epsfig}
\usepackage{graphicx}
\usepackage{amsmath}
\usepackage{amssymb}
\usepackage[utf8]{inputenc} 
\usepackage[T1]{fontenc}    
\usepackage{url}            
\usepackage{booktabs}       
\usepackage{amsfonts}       
\usepackage{nicefrac}       
\usepackage{microtype}      
\usepackage{graphicx}
\usepackage{multirow}
\usepackage{wrapfig}
\usepackage{xcolor}
\usepackage{bbm}
\usepackage{amsmath}
\usepackage{comment}
\usepackage{dsfont}
\usepackage{bbm}
\usepackage{wrapfig}
\usepackage{float}
\usepackage{enumitem}
\usepackage{amssymb}
\usepackage{stackengine}
\usepackage{stfloats}
\usepackage{capt-of}
\usepackage{cuted}
\usepackage{lipsum}
\def\delequal{\mathrel{\ensurestackMath{\stackon[1pt]{=}{\scriptstyle\Delta}}}}

\ifCLASSOPTIONcompsoc
  \usepackage[nocompress]{cite}
\else
  \usepackage{cite}
\fi
\ifCLASSINFOpdf
\else
\fi
\begin{document}
%
\title{DreamStone: Image as a Stepping Stone for Text-Guided 3D Shape Generation}
%
%
%
%

\author{Zhengzhe Liu,
        Peng Dai,
        Ruihui Li, Xiaojuan Qi, Chi-Wing Fu\\
        \
        \\
         \footnotesize{Project page: \url{https://liuzhengzhe.github.io/DreamStone.github.io/}}\\
\IEEEcompsocitemizethanks{\IEEEcompsocthanksitem Z. Liu and C.-W. Fu are with the Department
of Computer Science and Engineering, The Chinese University of Hong Kong, Hong Kong.\protect\\
E-mail: zzliu@cse.cuhk.edu.hk; cwfu@cse.cuhk.edu.hk.
\IEEEcompsocthanksitem P. Dai and X. Qi are with the Department of Electrical and Electronic Engineering, The University of Hong Kong, Hong Kong. \protect\\
E-mail: daipeng@eee.hku.hk; xjqi@eee.hku.hk. 
\IEEEcompsocthanksitem R. Li is with College of Computer Science and Electronic Engineering, The Hunan University, China.
Email: liruihui@hnu.edu.cn.
}
}

%
%

\markboth{IEEE TRANSACTIONS ON PATTERN ANALYSIS AND MACHINE INTELLIGENCE} 
{Shell \MakeLowercase{\textit{et al.}}: Bare Demo of IEEEtran.cls for Computer Society Journals}
%



\IEEEtitleabstractindextext{%
\begin{abstract}
This paper presents a new text-guided 3D shape generation approach DreamStone that uses images as a stepping stone to bridge the gap between the text and shape modalities for generating 3D shapes without requiring paired text and 3D data. 
The core of our approach is a {\em two-stage feature-space alignment strategy\/} that leverages a pre-trained single-view reconstruction (SVR) model to map CLIP features to shapes: to begin with, map the CLIP image feature to the detail-rich 3D shape space of the SVR model, then map the CLIP text feature to the 3D shape space through encouraging the CLIP-consistency between the rendered images and the input text. 
Besides, to extend beyond the generative capability of the SVR model, we design the text-guided 3D shape stylization module that can enhance the output shapes with novel structures and textures.
Further, we exploit pre-trained text-to-image diffusion models to enhance the generative diversity, fidelity, and stylization capability.
Our approach is generic, flexible, and scalable. It can be easily integrated with various SVR models to expand the generative space and improve the generative fidelity.
Extensive experimental results demonstrate that our approach outperforms the state-of-the-art methods in terms of generative quality and consistency with the input text. 
Codes and models are released at 
{\scriptsize \url{https://github.com/liuzhengzhe/DreamStone-ISS}}.
\end{abstract}

}

\maketitle

%
\IEEEpeerreviewmaketitle

\vspace*{-25pt}
\section{Introduction}\label{sec:intro}

3D shape generation has many practical applications, such as in CAD, 3D games, animations, and more. Among  different ways to generate 3D shapes, a user-friendly method is to generate shapes from text descriptions. This enables users to easily generate 3D shapes using natural language along with many applications in AR/VR and 3D printing. 
However, text-guided shape generation presents significant challenges owing to the difficulty of collecting paired text-shape data, the substantial semantic gap between texts and shapes, and the topological complexity of 3D shapes.


Previous research~\cite{chen2018text2shape,jahan2021semantics,liu2022towards} typically requires paired text-shape data for this challenging task. Yet, it is already non-trivial to collect 3D shapes, let alone manually annotate text-shape pairs, which incurs further complexities. Currently, the largest paired text-shape dataset available~\cite{chen2018text2shape} has only two categories, tables and chairs, significantly limiting the applicability of the existing works.

\begin{figure}
\centering
\includegraphics[width=0.99\columnwidth]{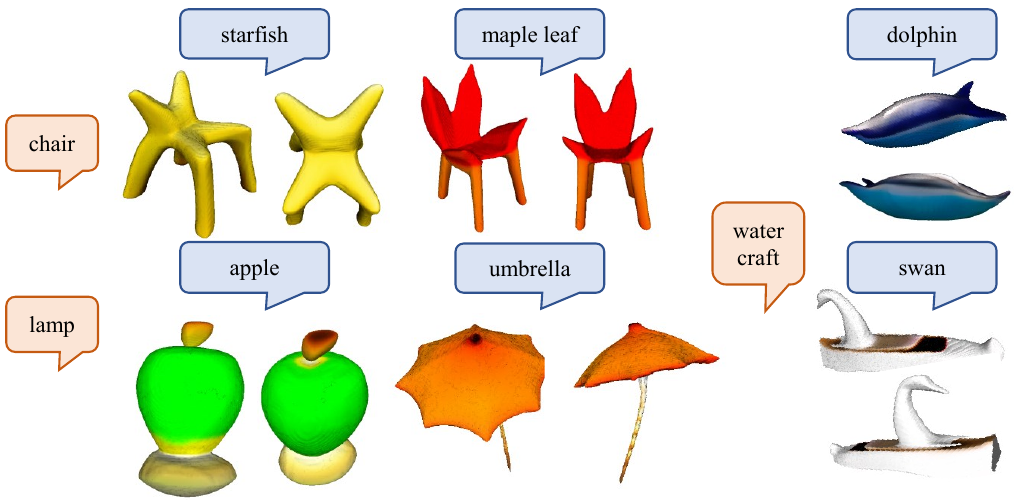}
\caption{Generative results of our DreamStone. The input text follows the prompt template ``A [shape] (in red boxes) imitates a [style] (in blue boxes)''. 
}
\label{fig:style_figure1}
\vspace*{-2.75mm}
\end{figure}

\begin{figure*}
\centering
\includegraphics[width=0.99\textwidth]{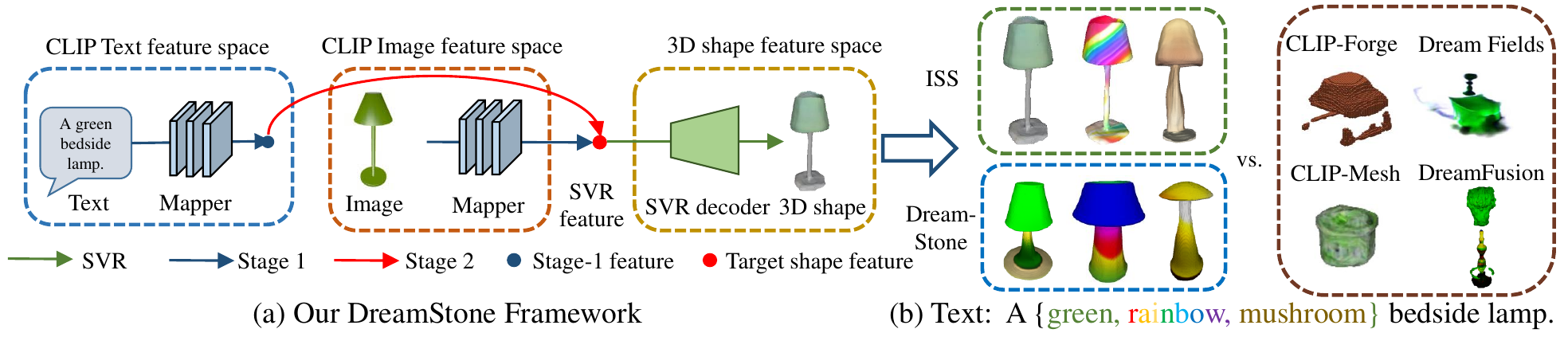}
\vspace*{-2mm}
\caption{The proposed DreamStone framework. 
Our two-stage feature-space alignment shown in (a) bridges the text space (the CLIP text feature) and the 3D shape space (the SVR feature) 
to generate 3D shapes from the text shown in (b), outperforming the existing works, 
without requiring paired text-shape data.
}
\label{fig:figure1}
\vspace*{-2.75mm}
\end{figure*}

Recently, several annotation-free approaches have been proposed for text-to-shape generation without requiring paired text-shape data. These approaches, such as CLIP-Forge~\cite{sanghi2021clip}, Dream Fields~\cite{jain2021zero}, CLIP-Mesh~\cite{mohammad2022clip}, and DreamFusion~\cite{poole2022dreamfusion}, utilize the large-scale language-vision models, e.g., CLIP~\cite{radford2021learning}, and text-to-image generation models, such as Imagen~\cite{saharia2022photorealistic}, for training. However, generating high-quality 3D shapes from unpaired text-shape data remains challenging for several reasons. First, due to the scarcity of 3D datasets, they can only generate a very limited range of shape categories. For instance, CLIP-Forge~\cite{sanghi2021clip} is struggling to generate shapes outside the ShapeNet dataset. Second, without injecting any text-related shape priors, it is also difficult to produce 
3D structures that match the input texts.  For example, CLIP Mesh~\cite{mohammad2022clip} and Dream Fields~\cite{jain2021zero} often generate 3D shapes incompatible with given texts (see Figure~\ref{fig:figure1} (b)) even with minutes or hours of test-time optimization for each shape instance. 
Third, the visual quality of the generated shapes is not satisfactory. As shown in Figure~\ref{fig:figure1} (b),  CLIP-Forge~\cite{sanghi2021clip} produces low-resolution outputs ({\ie}, $64^3$) without textures, the results generated by Dream Fields and CLIP-Mesh typically look surrealistic (rather than real), and the 3D topology and surface quality of DreamFusion still have a large room for improvement.

Going beyond existing approaches, we present a novel text-guided 3D shape generation method without requiring paired text-shape data. We propose to {\em leverage 2D Image as a Stepping Stone \/}  to implicitly bridge the shape and text modalities and exploit diffusion models for enhanced diversity, quality, and generative scope, namely DreamStone. Specifically, we employ the pre-trained vision-language model CLIP to train a mapper that maps CLIP image features to a pre-trained 3D shape space. In inference, this mapper maps the CLIP text features to the target shape space, as shown in Figure~\ref{fig:figure1} (a) stage 1. However, there exists a gap between the CLIP image and text features. As a result, the CLIP text feature might not be mapped to a desired shape feature. To tackle this issue, we further fine-tune the mapper to improve the text-shape consistency. We do this by adopting a training objective encouraging CLIP consistency between the input text description and rendered images. This fine-tuning stage is depicted in Figure~\ref{fig:figure1} (a) as stage 2. Also, text-guided 3D shape generation is a one-to-many mapping problem, i.e., a single input text can correspond to multiple 3D shapes. To enhance the generative diversity of the two-stage feature-space alignment, we employ an off-the-shelf diffusion model, which is referred to as “diffusion prior” in this paper, to map the CLIP text feature to CLIP image feature and sample multiple generated CLIP image features that match the text feature  to produce diverse results at inference. The two-stage feature-space alignment can generate plausible shapes from texts.

To go beyond the generative space of the pre-trained SVR models, we design CLIP-guided shape stylization and Score Distillation Sampling (SDS)-guided refinement modules that enable the generation of new and visually pleasing textures and structures during testing. Specifically, the CLIP-guided shape stylization module updates the decoder of the SVR model by optimizing the CLIP consistency between the rendered images from the generated shapes and the target style descriptions. 
Though this strategy help expands the models' generative capability toward open-world style descriptions, it suffers from generating local detailed structure due to the global guidance of CLIP features; see Figure~\ref{fig:figure1} ``ISS''.
Hence, to produce fine-grained structures and high-fidelity textures, we explore leveraging pre-trained diffusion models and marry Score Distillation Sampling (SDS)~\cite{poole2022dreamfusion} with our two-stage feature-space alignment framework. This involves utilizing SDS to provide a loss function for updating our decoder.
This allows us to generate high-fidelity novel structures and  textures and even create imaginary shapes by incorporating the semantic
attributes of the target style into the shape; see Figure~\ref{fig:style_figure1} and Figure~\ref{fig:figure1} ``DreamStone''.  
This also extends the generation capability of our DreamStone to unseen categories out of the image dataset. 
 Besides, by leveraging the 3D shape prior of the two-stage feature-space alignment, our model outperforms \cite{poole2022dreamfusion} in terms of surface quality and topology faithfulness, while typically requiring much fewer training iterations.

Lastly, our approach can be compatible with various SVR models~\cite{niemeyer2020differentiable,alwala2022pre,gao2022get3d}.
For instance, 
we can adopt SS3D~\cite{alwala2022pre} to generate shapes using single-view in-the-wild images, thus expanding our approach's generative capability beyond the 13 categories of ShapeNet that can be generated by~\cite{sanghi2021clip}.
Also, our approach can work with the very recent method GET3D~\cite{gao2022get3d} to generate
high-quality 3D shapes from text; see results in Section~\ref{sec:results}.

In summary, our approach expands the boundary of 3D shape generation from texts in the following aspects. First, we cast the challenging text-guided shape generation task to be a single-view reconstruction (SVR) task, which is more approachable. 
Second, our approach is efficient. It can create plausible 3D shapes in only 85 sec. with the two-stage feature-space alignment and high-quality and stylized 3D shapes with Score Distillation Sampling in less than 30 min. vs. 72 min. of Dream Fields~\cite{jain2021zero} and 90 min. of DreamFusion~\cite{poole2022dreamfusion} (using the Stable-Dreamfusion version due to the lack of public code)\footnote{We use the latest version of an available public implementation of DreamFusion, Stable-Dreamfusion~\cite{stable-dreamfusion}, with the commit ``099468e6'' updated on Feb 7, 2023, as the official code of DreamFusion has not been released.}. Further, the generation capabilities of our approach outperform the generation capabilities of the state-of-the-art approaches; see Figure~\ref{fig:figure1} (b). Lastly, our approach is generic, scalable, and compatible with a wide range of SVR methods.

\vspace*{1mm}
\noindent \textbf{Different from Our Conference Paper.} \ This manuscript extends ISS~\cite{liu2023iss}, a spotlight paper at International Conference on Learning Representations 2023.  Particularly, this extended version addresses several limitations of the conference version. First, although~\cite{liu2023iss} can create diverse results given one text prompt, it does not ensure that all the results are consistent with the text. Moreover, the generation quality of~\cite{liu2023iss} is not very high due to the lack of fine details. To address these issues, we extend ISS~\cite{liu2023iss} in the following aspects. First, we extend ISS with a diffusion prior~\cite{ramesh2022hierarchical} to generate more diversified 3D shapes while ensuring their consistency with the given text. Then, we propose an SDS-guided refinement module to further improve the fidelity of the generated shapes. Further than that, our SDS-guided stylization enables the generation of imaginary 3D shapes complementing our previous CLIP-guided stylization~\cite{liu2023iss}.  Last, we conduct more experiments on 3D shape generation and shape stylization, and compare DreamStone with the latest works CLIP-Mesh~\cite{mohammad2022clip} and DreamFusion~\cite{poole2022dreamfusion}. Our experimental results, both quantitative and qualitative, demonstrate that our approach is able to surpass the state-of-the-art methods in text-guided 3D shape generation.
\section{Related Works}





\textbf{Text-Guided Image Generation.}
Text-guided image synthesis has been intensively studied recently~\cite{reed2016generative,reed2016learning,zhang2017stackgan,zhang2018stackgan++,xu2018attngan,li2019controllable,li2020manigan,qiao2019mirrorgan,wang2021cycle,stap2020conditional,yuan2019bridge,souza2020efficient,wang2020text,rombach2020network,patashnik2021styleclip,xia2021tedigan}.
Leveraging auto-regressive and diffusion models, recent works achieve impressive performance on text-guided image generation~\cite{ramesh2021zero,ding2021cogview,nichol2021glide,liu2021fusedream,ramesh2022hierarchical,saharia2022photorealistic} to produce images of many classes. To avoid the need for text data, Wang et al.~\cite{wang2022clip} and Zhou et al.~\cite{zhou2021lafite} explore text-free text-to-image generation leveraging CLIP.

Beyond text-guided image generation, it is more challenging to create 3D shapes from text. First, unlike paired text-image data that can be readily collected from the Internet, it is laborious and challenging to acquire large paired text-shape data. Second, text-to-shape generation aims to predict complete 3D structures beyond a single 2D view in text-guided image generation. Third, there are more complex spatial structures and topologies in 3D shapes beyond 2D images in regular pixel grids, making it even more challenging to generate 3D shapes from texts. 


\vspace*{3mm}
\noindent\textbf{Text-Guided 3D Shape Generation.}
In this research field, some approaches require paired text-shape data, such as \cite{chen2018text2shape,jahan2021semantics,liu2022towards}. However, to avoid the need for paired data, recent works such as CLIP-Forge\cite{sanghi2021clip}, Dream Fields~\cite{jain2021zero}, CLIP-Mesh~\cite{mohammad2022clip}, and DreamFusion~\cite{poole2022dreamfusion} leverage pre-trained vision-language models or text-to-image models. Despite their advancements, these approaches still have limitations, as discussed in Section~\ref{sec:intro}. Moreover, some works use CLIP to manipulate 3D shapes/NeRF using text~\cite{michel2021text2mesh, jetchev2021clipmatrix, wang2021clipnerf, chen2022tango} and generate 3D avatars~\cite{hong2022avatarclip}. In contrast, our approach presents a new framework for text-guided 3D shape generation without the need for paired text-shape data, using the newly proposed two-stage feature-space alignment. Our experimental results demonstrate superior fidelity and text-shape consistency beyond existing methods.

\vspace*{3mm}
\noindent\textbf{Differentiable Rendering}. 
As a powerful technique, differentiable rendering enables 3D models to be optimized using 2D images. There are numerous applications, such as generating 3D shapes from 2D images or reconstructing 3D objects from multiple 2D views. By modeling the rendering process as a differentiable function, gradients can be computed with respect to the input parameters of the function, allowing for efficient optimization using gradient-based techniques. This has led to significant advances in fields such as computer vision and computer graphics. Recent works~\cite{niemeyer2020differentiable,munkberg2022extracting,gao2022get3d} leverage differentiable rendering for 3D shape generation using 2D images. In this work, we derive 2D images of the generated 3D shape using differentiable rendering and use a pre-trained large-scale image-language model CLIP to encourage consistency between 2D images and input texts. Thanks to differentiable rendering, we can update the generated 3D shapes indirectly using the rendered images. 

\vspace*{3mm}
\noindent\textbf{Single-View Reconstruction (SVR).} 
%
This work is also related to SVR. SVR has recently been explored with voxels~\cite{zubic2021effective}, meshes~\cite{agarwal2020gamesh}, and implicit fields~\cite{niemeyer2020differentiable,alwala2022pre}. 
In this work, we leverage an SVR model to bridge the image and shape modalities, thus allowing us to use 2D images as a stepping stone to produce 3D shapes from texts. Moreover, our approach is flexible since we map the features in the latent space implicitly rather than explicitly.
\vspace*{-5pt}
\section{Methodology}

\begin{figure*}
\centering
\includegraphics[width=0.99\textwidth]{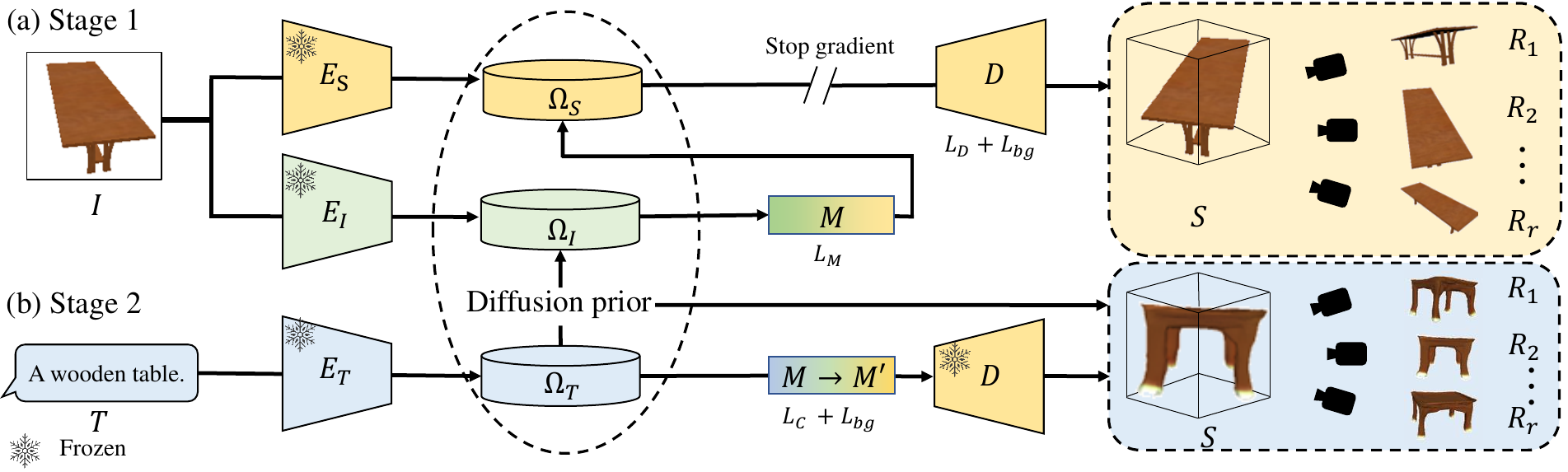}
\vspace*{-2mm}
\caption{
Overview of our two-stage feature-space alignment. 
(a) In the first stage, we align the CLIP image feature space $\Omega_{\text{I}}$ and the shape space $\Omega_{\text{S}}$ of a pre-trained single-view reconstruction (SVR) model with a CLIP2Shape mapper $M$, which maps images to shapes while keeping $E_\text{S}$ and $E_\text{I}$ frozen. Then we fine-tune the decoder $D$ using $L_{\text{bg}}$ to encourage the background color to be white. During training, we stop gradients of the SVR loss $L_D$ and the background loss $L_{\text{bg}}$ to eliminate their effects on $M$.
(b) In the second stage, we introduce a fast-time optimization by fixing the decoder $D$ and fine-tuning the mapper $M$ to $M'$, further encouraging the CLIP consistency between the rendered images of the generated shape and the input text $T$.
}
\label{fig:overview}
\vspace*{-2.5mm}
\end{figure*}

\subsection{Overview}


To generate 3D shape $S$ from text $T$ without relying on paired text-shape data, we map the CLIP features to a latent shape feature space of a pre-trained SVR model, leveraging the joint text-image feature embeddings from CLIP and also the 3D shape prior learned by the SVR model. 
Here, we leverage multi-view RGB/RGBD images and the corresponding camera poses for training, without needing the paired text-shape data.
The framework has four components: (1) image encoder $E_{\text{S}}$ to map the input image $I$ to shape space $\Omega_{\text{S}}$ of the SVR model, (2) pre-trained CLIP image and text encoders $E_{\text{I}}$ and $E_{\text{T}}$ that map image $I$ and text $T$ to CLIP feature spaces $\Omega_{\text{I}}$ and $\Omega_{\text{T}}$, (3) mapper $M$ consisting of 12 fully-connected and Leaky-ReLU layers to map CLIP image features to the latent shape space $\Omega_{\text{S}}$ of SVR, and (4) decoder $D$ that generates the 3D shape $S$. The proposed approach uses DVR~\cite{niemeyer2020differentiable} as the SVR model in the experiments unless specified otherwise.

Generally speaking, we present a novel two-stage feature-space alignment approach to bridge the image, text, and shape modalities. First, we train the mapper $M$ to bridge the CLIP image space $\Omega_\text{I}$ and the shape space $\Omega_\text{S}$, as shown in Figure~\ref{fig:overview}(a). Afterward, at test time, $M$ is fine-tuned to further bridge the CLIP text space $\Omega_\text{T}$ and $\Omega_\text{S}$, as shown in Figure~\ref{fig:overview}(b). Finally, we can optionally improve the texture and structure generation capability of our model by fine-tuning the decoder $D$ (as shown in Figure~\ref{fig:overview_style}). 

In Section~\ref{sec:empirical}, we begin by presenting two empirical studies that investigate the properties of the CLIP feature space. We then introduce our two-stage feature-space alignment approach in Section~\ref{sec:two-sgage}. Following that, in Section~\ref{sec:stylization}, we present our method for text-guided shape refinement and stylization. Finally, in Section~\ref{sec:compatible}, we discuss that our approach is compatible with different SVR models and how we can extend our method to generate a wide range of categories and high-quality shapes.

\vspace*{-3pt}
\subsection{Empirical Studies and Motivations}\label{sec:empirical}


Prior works on text-guided 3D shape generation mainly use CLIP without analyzing its workings and limitations. To gain a better understanding of the CLIP feature space and its suitability for text-guided 3D shape generation, we conduct two empirical studies.

\begin{figure*}
\centering
\includegraphics[width=0.99\textwidth]{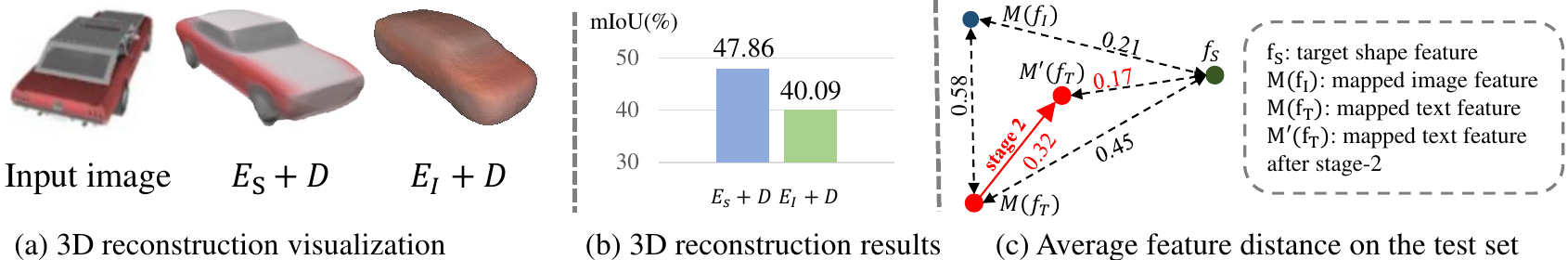}
\vspace*{-1.5mm}
\caption{Results of empirical studies on CLIP feature spaces. 
} 
\label{fig:motivation}
\vspace*{-2.5mm}
\end{figure*}

\vspace*{-3pt}
\subsubsection{Whether the CLIP feature is suitable for 3D shape generation?}

In the first empirical study, we investigate whether the CLIP image feature space $\Omega_\text{I}$ has enough representative capability for 3D shape generation by attempting to generate shapes from $\Omega_\text{I}$. 
To do so, we train the SVR model by adopting the CLIP image encoder $E_\text{I}$ to replace the original SVR image encoder $E_{\text{S}}$. At the same time, we optimize the decoder $D$ using the same loss function as DVR~\cite{niemeyer2020differentiable} with $E_{\text{I}}$ frozen.
This design is inspired by the motivation that we can generate 3D shapes from the text by adopting the CLIP text encoder $E_\text{T}$ to replace $E_\text{I}$ in inference. 
To evaluate the 3D shape generative capability $E_\text{S}$ and $E_\text{I}$, we measure the 3D mIoU of their generated shapes and ground truths (Figure~\ref{fig:motivation} (b)).
The result indicates that the representative capability of CLIP image encoder $E_\text{I}$ is inferior to $E_\text{S}$ due to its inferior capability to capture input image details that are necessary for 3D shape generation. 
This result is easy to understand since CLIP image encoder $E_\text{I}$ has been optimized to extract semantic-aligned features with the paired text data in the training of CLIP, instead of being encouraged to capture image details. 
As a result in Figure~\ref{fig:motivation} (a), image details that are necessary for 3D reconstruction, such as textures, are overlooked by $E_\text{I}$. In contrast, $E_\text{S}$ in the SVR model is trained for 3D generation and is encouraged to capture the necessary image details. 
These results indicate that we can generate shapes from $\Omega_\text{S}$ instead of $\Omega_\text{I}$ to improve the generative quality. To do so, we design a mapper $M$ to map from CLIP image feature space $\Omega_\text{I}$ to shape space $\Omega_\text{S}$ to enable the generation from $\Omega_\text{S}$.

\vspace*{-3pt}
\subsubsection{Does the CLIP image and text feature gap affects 3D shape generation?} \label{sec:empirical2}

The second investigation aims to analyze the gap between the normalized CLIP text feature $f_\text{T}\in \Omega_\text{T}$ and image feature $f_\text{I} \in \Omega_\text{I}$ as shown in Figure~\ref{fig:figure1} (a) and examine how this gap affects text-guided 3D shape generation. 
Specifically, we measure the cosine distance between $f_\text{I}$ and $f_\text{T}$ based on the text and rendered images of the randomly selected 300 text-shape pairs from the text-shape dataset~\cite{chen2018text2shape} as follows:
\begin{equation}
d=1-\text{cosine\_similarity}(f_\text{I},f_\text{T}).
\label{equ:clip-d}
\end{equation}
The result $d(f_\text{T},f_\text{I})=0.783\pm0.004$ through three repetitions of the experiment suggests that there is still a certain gap between the paired text and image features. Additionally, the angle between the two features is around $\arccos(1-0.783)=1.35$ radians in this text-shape dataset~\cite{chen2018text2shape}. The above result implies that the generated 3D shape may not be consistent with the input text if we simply replace $f_\text{I}$ with $f_\text{T}$ in inference. As demonstrated in Figure~\ref{fig:motivation} (c), this simple strategy results in a cosine distance 0.45 to $f_\text{S}\in \Omega_\text{S}$, much larger than $d(M(f_\text{I}), f_\text{S})=0.21$. This finding is consistent with the results reported in~\cite{liang2022mind}. To address this issue, we propose to fine-tune $M$ into $M'$ at test time, aiming at producing a feature $M'(f_\text{T})$ that has a smaller distance to $f_\text{S}$ compared with $M(f_\text{T})$.

\vspace*{-5pt}

\subsection{Two-Stage Feature-Space Alignment}\label{sec:two-sgage}

Based on these findings, we propose a two-stage feature-space alignment approach that connects the image space $\Omega_\text{I}$ and shape space $\Omega_{\text{S}}$ in the first stage and further connects the text space $\Omega_\text{T}$ to shape space $\Omega_{\text{S}}$ in the second stage, with the image space $\Omega_\text{I}$ as a stepping stone. 

\vspace*{-3pt}
\subsubsection{Stage-1: CLIP image-to-shape alignment}

Figure~\ref{fig:overview} (a) illustrates the stage-1 alignment. On the one hand, the shape space $\Omega_{\text{S}}$ is able to capture richer image details compared with CLIP image space $\Omega_{\text{I}}$. On the other hand, $\Omega_{\text{I}}$ helps to enable the text input thanks to its joint text-image embedding with $\Omega_{\text{T}}$. Inspired by the above two motivations,
we design a CLIP2Shape mapper $M$ consisting of $12$ fully-connected layers to map $f_{\text{I}}$ to $\Omega_{\text{S}}$. To optimize $M$, we use $L_2$ regression loss between the mapped CLIP image feature $M(f_{\text{I}})$ and pre-trained SVR feature encoder $f_\text{{S}}=E_{\text{S}}(I)$ as shown in Equation~(\ref{equ:stage-1}):
\begin{equation}
\begin{aligned}
\mathcal{L}_{M}=\sum^N_{i=1} ||E_{\text{S}}(I_i)-M(f_{\text{I},i})||^2_2 
\label{equ:stage-1}
\end{aligned}
\end{equation}
where $N$ and $f_{\text{I},i}$ indicates the total number of images for training and the normalized CLIP feature of $I_i$, respectively.

In addition, we incorporate a fine-tuning module for decoder $D$ to encourage it to generate 3D shapes with a white background. This module helps the model to focus on object-centric features while ignoring the background (see Figure~\ref{fig:bg}). Specifically, we propose a novel background loss $\mathcal{L}_\text{{bg}}$ in Equation~(\ref{equ:bg}) below, which enhances the model's ability to capture foreground objects and prepares it for the second-stage alignment.

\begin{equation}
\begin{aligned}
\mathcal{L}_\text{bg}=\sum_p ||D_c(p)-1||_2^2 \mathbbm{1}(F\cap\text{ray}(o,p) =\emptyset)
\label{equ:bg}
\end{aligned}
\end{equation}
where $p$ means a query point coordinate, $D_o(p)$ and $D_c(p)$ are the occupancy and color prediction of $p$, respectively. $F=\{p: D_o(p)>t\}$ indicates the object region, where $D_o(p)$ is greater than a pre-defined threshold $t$. $F\cap\text{ray}(o,p) =\emptyset$ means the background region where a ray connecting camera center $o$ and $p$ does not intersect the object. Besides, $\mathbbm{1}$ is the indicator function and $\mathbbm{1}(F\cap\text{ray}(o,p) =\emptyset)=1$ if $p$ is in the background region. 
To summarize, $\mathcal{L}_{\text{bg}}$ is designed to encourage the background region to be predicted as the white color (value 1) and assist the model in better capturing the generated shape. Besides, the same set of loss functions $\mathcal{L}_{\text{D}}$ from DVR~\cite{niemeyer2020differentiable} is still adopted for maintaining the capability to generate 3D shapes of the SVR model.

Hence, the total loss in stage 1 is $\lambda_{M}\mathcal{L}_{M}$ for mapper $M$ and $\lambda_{\text{bg}}\mathcal{L}_{\text{bg}}+\mathcal{L}_{D}$ for decoder $D$, where $\lambda_{\text{bg}}$ and $\lambda_{M}$ indicate loss weights. 
The stage-1 alignment is trained with multi-view RGB/RGBD images and provides a good starting point for the stage 2 per-text optimization.

\vspace*{-3pt}
\subsubsection{Stage-2: text-to-shape alignment}


After bridging the image and shape modalities, we further try to bridge the text and shape modalities by proposing a fast test-time optimization in stage 2 to reduce the gap between the CLIP features of the input text $T$ and image $I$, as discussed in the second empirical study. By doing so, we can encourage the generated shape $S$ to be more consistent with the input text. As we cannot directly optimize the similarity between the text and shape features, reducing the semantic gap between $f_\text{T}$ and $f_\text{I}$ is an effective way to align the two modalities and improve the overall performance.


As illustrated in Figure~\ref{fig:overview} (b), the stage-2 alignment starts by replacing $E_{\text{I}}$ with $E_{\text{T}}$ to extract CLIP text feature $f_{\text{T}}$, given the input text $T$. We then fine-tune the mapper $M$ using a CLIP consistency loss to reduce the gap between the input text $T$ and $m$ rendered images $\{R_i\}_{i=1}^m$ captured from random camera viewpoints of the output shape $S$. The CLIP consistency loss is defined in Equation~\eqref{equ:stage-2}. By minimizing this loss, we encourage the output shape to be consistent with the input text.

\begin{equation}
\begin{aligned}
\mathcal{L}_{\text{C}}=\sum_{i=1}^m{\langle{f_\text{T}} \cdot \frac{E_{\text{I}}(R_i)}{||E_{\text{I}}(R_i)||}\rangle} 
\label{equ:stage-2}
\end{aligned}
\end{equation}
where $\langle\cdot\rangle$ indicates the inner-product. 

\begin{figure*}
\centering
\includegraphics[width=0.99\textwidth]{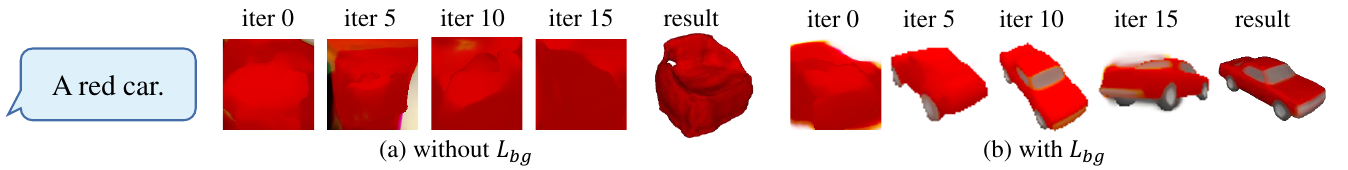}
\vspace*{-1.5mm}
\caption{Generating shapes from text with and without our background loss $\mathcal{L}_\text{bg}$. Input text: A red car. 
}
\label{fig:bg}
\vspace*{-2.5mm}
\end{figure*}


In stage-2 alignment, we continue to use $\mathcal{L}_{\text{bg}}$ to improve the model's object awareness. Figures~\ref{fig:bg} (a) and (b) indicate that the model can find a rough shape that fits the input text in about five iterations when $\mathcal{L}_\text{{bg}}$ is used. On the other hand, without $\mathcal{L}_\text{{bg}}$, the model fails to produce a reasonable output because the same color predicted in both the object and background regions impedes the model's ability to perceive the object.


Our stage-1 alignment has already narrowed the semantic gap between text space $\Omega_\text{T}$ and shape space $\Omega_\text{S}$ with $M$.
So, the stage-2 alignment just requires fine-tuning $M$ using a CLIP consistency loss with the input text for only 20 iterations. This fine tuning takes around 85 seconds on one GeForce RTX 3090 Ti GPU, which is significantly faster than Dream Fields~\cite{jain2021zero} (72 minutes) and DreamFusion~\cite{poole2022dreamfusion} (90 minutes). After stage-2 alignment, a plausible result can be obtained readily, shown as ``result'' in Figure~\ref{fig:bg} (b). Our two-stage feature-space alignment is a new approach that can efficiently generate 3D shapes from text, significantly reducing the test time compared to prior methods.

\vspace*{-3pt}
\subsubsection{Diversified 3D shape generation with diffusion prior}

In general, 3D shape generation from text is a one-to-many task, meaning that multiple plausible shapes can correspond to the same piece of text. 
To account for this, instead of using a single objective using $f_\text{T}$ to construct $\mathcal{L}_\text{C}$, we propose to sample features from a pre-trained text-to-image diffusion model~\cite{ramesh2022hierarchical}, which can generate features $f_{\text{T}\rightarrow\text{I}}$ in the CLIP image feature space from a single input text CLIP feature $f_{\text{T}}$. 
At each time, to generate a 3D shape, we  obtain one text-to-image feature by sampling a random noise and obtain $f_{\text{T}\rightarrow\text{I}}$. 
This $f_{\text{T}\rightarrow\text{I}}$ is then combined with the original text feature $f_\text{T}$ to construct $\mathcal{L}_C$  in the two-stage feature alignment as 
\begin{equation}
\begin{aligned}
\mathcal{L}_{\text{C}}=\sum_{i=1}^m{\langle{ (\tau f_{\text{T}\rightarrow\text{I}}+(1-\tau) f_\text{T}}) \cdot \frac{E_{\text{I}}(R_i)}{||E_{\text{I}}(R_i)||}\rangle}.  
\label{equ:div}
\end{aligned}
\end{equation}
where $f_{\text{T}\rightarrow\text{I}}$ is the predicted $f_\text{I}$ from $f_\text{T}$ by diffusion prior~\cite{ramesh2022hierarchical} with sampled random noise and $\tau$ is a hyperparameter that balances diversity and text-shape consistency; a larger $\tau$ leads to more diverse shapes, while a smaller $\tau$ encourages more consistency between the text and shape.

By sampling multiple random noises which deliver multiple $f_{\text{T}\rightarrow\text{I}}$ and constructing different consistency objective $\mathcal{L}_C$, our model can be optimized to generate diverse results at the test time; see Figure~\ref{fig:overview} (b) ``diffusion prior''. This allows our model to create diverse 3D shapes for the same piece of input text.

Our approach is inspired by the ``diffusion prior'' method, which generates a CLIP image feature from the CLIP text feature described in~\cite{ramesh2022hierarchical}. We use the same name 
``diffusion prior'' to maintain consistency.
It is important to note that the diffusion prior module is a highly efficient technique for diversified generation. Instead of producing a complete image, the diffusion prior generates an image feature vector $f_{T\rightarrow I}$, conditioned on a text feature vector $f_T$ in the latent space. This process takes only $0.78\pm0.09$ seconds on an RTX 3090 Ti.
Furthermore, to generate a 3D shape, the aforementioned diffusion process is conducted only once. Specifically, given text $T$, we employ the diffusion process to create an associated image feature $f_{T\rightarrow I}$, requiring only 0.78 seconds. We then perform our stage-2 alignment using Equation~\ref{equ:div} to generate a 3D shape, taking approximately 85 seconds. Consequently, the total time remains around 85 seconds, and the diffusion module does not introduce significant computational overhead.
In summary, a single diffusion prior process can produce a unique $f_{T\rightarrow I}$ (0.78 seconds), followed by the creation of a unique 3D shape (85 seconds). Multiple iterations of this process enable diversified generation due to the randomness of the diffusion prior, with each generation requiring roughly $0.78+85$ seconds.

Besides, by exploiting the prior diffusion models, our model can also better mitigate the effect of the semantic gap between $f_\text{T}$ and $f_\text{I}$ in the stage-2 alignment; see the discussion in Section~\ref{sec:empirical2}.  This is achieved by encouraging $f_\text{I}$ of the rendered images to be consistent with the blended features of the sampled text-to-image feature $f_{\text{T}\rightarrow\text{I}}$ and the input text $f_\text{T}$, rather than just the input text feature $f_\text{T}$ itself.

\vspace*{-3pt}
\subsection{Text-Guided 3D Shape Stylization}\label{sec:stylization}

\begin{figure*}
\centering
\includegraphics[width=0.99\textwidth]{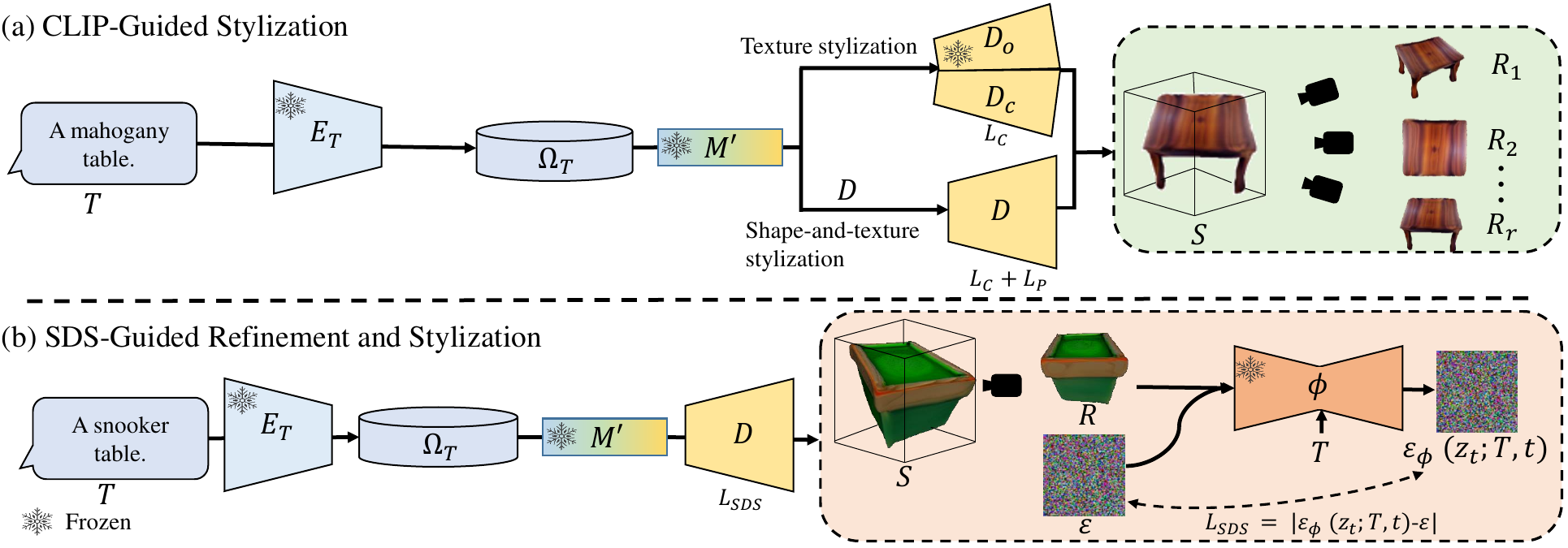}
\vspace*{-2mm}
\caption{
Our text-guided 3D shape refinement and stylization framework. 
(a) CLIP-guided stylization. (b) SDS-guided refinement and stylization. 
}
\label{fig:overview_style}
\vspace*{-2.5mm}
\end{figure*}


While the two-stage feature-space alignment can generate plausible 3D shapes as shown in Figures~\ref{fig:overview} (b) and~\ref{fig:bg} (b), its generative space and quality are still limited by the pre-trained SVR model in use. For instance, DVR~\cite{niemeyer2020differentiable} cannot generate shapes beyond the synthetic patterns in ShapeNet dataset.  
Further, to enable the model to generate a broader range of structures and textures, we introduce text-guided stylization and refinement modules to enable our approach to create shapes out of the SVR generative space with delicate structures and textures; see Figures~\ref{fig:overview_style} and~\ref{fig:figure1} ``DreamStone''.

\subsubsection{CLIP-guided stylization}\label{sec:clip_stylization}

First, we introduce CLIP-guided stylization to stylize 3D shapes beyond the generative space of the adopted SVR model. 

The top branch of Figure~\ref{fig:overview_style} (a) shows how we apply this method for texture stylization. To begin with, we duplicate $D$, except for the output layer, to create two networks: $D_o$ for occupancy prediction and $D_c$ for color prediction. Then we decompose the output layer to be $1$ and $3$ channels for occupancy and color prediction, respectively, and place them on the top of $D_o$ and $D_c$.  

To further create new structures for shape stylization, we incorporate a shape-and-texture stylization strategy in addition to texture stylization, as depicted in the bottom branch of Figure~\ref{fig:overview_style} (a). To do so, we further optimize $D$ by adopting the CLIP consistency loss in Equation~\ref{equ:stage-2}. Besides, to preserve 3D prior learned in the two-stage feature-space alignment, we additionally propose a 3D prior loss $\mathcal{L}_P$ as shown in Equation~(\ref{equ:prior}). 
\begin{equation}
\mathcal{L}_\text{P}=\sum_p |D_o(p)-D_o^\prime(p)|
\label{equ:prior}
\end{equation}
where $D_o(p)$, $D_o^\prime(p)$ indicate the initial occupancy prediction from $D$ and the optimized $D$ in the stylization training process of the query point $p$, respectively.



\begin{figure*}
\centering
\includegraphics[width=0.99\textwidth]{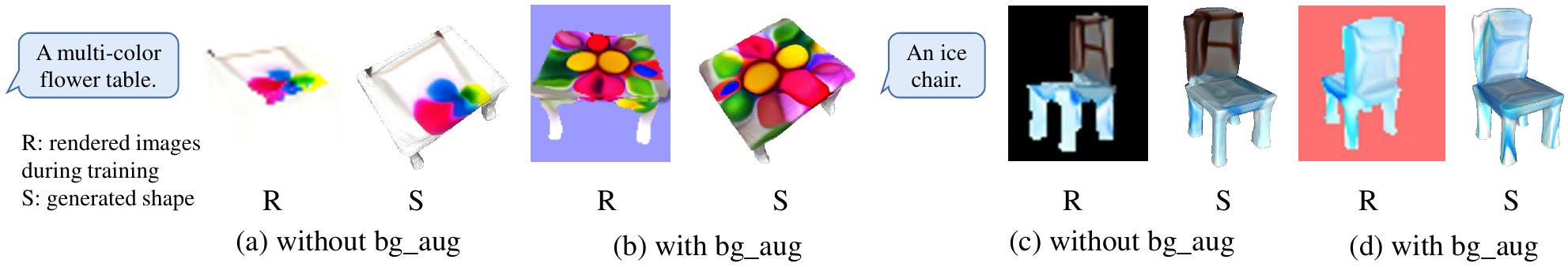}
\vspace*{-1.5mm}
\caption{Text-guided 3D shape stylization with and without our background augmentation.}
\label{fig:bg_aug_supp}
\vspace*{-2.5mm}
\end{figure*}

To enhance the network's object awareness in the stylization process, we introduce a background augmentation technique. 
As illustrated in Figure~\ref{fig:bg_aug_supp} (a), when the shape is in white, it can blend into the white background, making it difficult for the model to capture the object boundaries and resulting in textures that are poorly aligned with the table. Similarly, in Figure~\ref{fig:bg_aug_supp} (c), the generated texture is adversely harmed by the background color which is black, leading to inferior stylization results.
In our background augmentation strategy, 
we propose to substitute the background color as a random RGB value for each training iteration. In this way, the object region is easily distinguishable during training, as depicted in Figure~\ref{fig:bg_aug_supp} (b, d), leading to an improvement in texture-shape consistency and stylization quality.


\subsubsection{SDS-guided refinement and stylization}\label{sec:sds_stylization}

The CLIP-guided stylization 
helps generate
3D shapes outside the scope of the SVR model's generative space. Yet, the quality of the generated shapes is still bounded by the adopted SVR model {with detailed structures missing}. To further enhance the quality of the generated shapes, we introduce a new SDS-guided refinement and stylization technique 
to decorate the 3D shapes with intricate details and textures, as shown in Figures~\ref{fig:style_figure1},~\ref{fig:figure1} ``DreamStone'', and Figures~\ref{fig:overview_style} (b).


The proposed SDS-guided refinement module is inspired by~\cite{poole2022dreamfusion} and aims to improve the generative quality of the pre-trained SVR model.
Given a pre-trained text-guided image generation diffusion model $\phi$ and an input text $T$, we adopt Score Distillation Sampling (SDS) approach to fine-tune $D$ by encouraging the rendered image $R$ to be closer to the generated image of $\phi$ given input $T$. Specifically, we adopt Stable-Diffusion~\cite{rombach2022high} as $\phi$.
%
As shown in Figure~\ref{fig:overview_style} (b), we use $\theta$ to denote parameters in the decoder $D(p; \theta)$, $p$ to represent the query points, and $R(D(p; \theta))$ to indicate the rendered image from a randomly chosen viewpoint. 
Specifically, we randomly sample a time step $t$ and add noise to $R$ to produce $z_t$: $z_t=\sqrt{\bar{\alpha}_t} R+\sqrt{1-\bar{\alpha}_t} \epsilon$. 
The text $T$ and $z_t$ are fed into the pre-trained diffusion model which predicts the noise $\hat{\epsilon}_\phi(z_t; T, t)$. 
The predicted noise is compared with the added noise $\epsilon$ to construct the $\mathcal{L}_\text{sds}$.  
The procedure for calculating the gradient $\nabla_\theta \mathcal{L}_\text{sds}$ 
is illustrated below. 

\begin{subequations}
\begin{align}
&\nabla_\theta \mathcal{L}_\text{sds}(\phi, R(D(p;\theta)))=
\mathbb{E}_{t,\epsilon}[\frac{\partial (\hat{\epsilon}_\phi(z_t; T, t)-\epsilon)}{\partial \theta}] \label{equ:sdsa}\\
&=\mathbb{E}_{t,\epsilon}[(\hat{\epsilon}_\phi(z_t; T, t)-\epsilon)\frac{\partial \hat{\epsilon}_\phi(z_t; T, t)}{\partial z_t}\frac{\partial z_t}{\partial R}\frac{\partial R}{\partial \theta}] \label{equ:sdsb}\\
&\delequal\mathbb{E}_{t,\epsilon}[w(t)(\hat{\epsilon}_\phi(z_t; T, t)-\epsilon)\frac{\partial R}{\partial \theta}] \label{equ:sdsc}
\end{align}
\end{subequations}
where $w(t)={\partial z_t}/{\partial R}=\sqrt{\bar{\alpha}_t} I$ is a weighting function, and the term $\frac{\partial \hat{\epsilon}_\phi(z_t; T, t)}{\partial z_t}$ can be omitted indicated by~\cite{poole2022dreamfusion}. 
{The gradient $\nabla_\theta \mathcal{L}_\text{sds}$ will encourage the parameters $\theta$ to be updated so that the model can produce rendered images $R$ moving toward the high-density region of the score function. This means that the rendered image $R$ will be encouraged to be realistic and match the text, which in turn will help update the parameters $\theta$. 
}


The SDS-guided refinement further enhances the surface details of the generated shapes while preserving the overall topology learned by the two-stage feature-space alignment. With much fewer
training iterations than DreamFusion, DreamStone is able to generate 3D shapes with comparable or even higher fidelity, as shown in Section~\ref{sec:results}. Besides, our DreamStone helps mitigate the ``multi-face Janus problem'' of DreamFusion~\cite{poole2022dreamfusion}, which refers to the situation that the generated shapes have multiple, often disconnected, faces. This can occur due to the lack of constraints on the topology of the generated shapes without 3D priors. 
In contrast, our DreamStone leverages the 3D shape prior learned in the two-stage feature-space alignment to encourage consistency in the shape topology and achieve faithful and coherent shapes. 
Further, SDS enables DreamStone to generate a broader 
range of 3D shapes out of the image dataset; we will provide more results in Section~\ref{sec:results}.

\textbf{Discussions on stage-2 alignment and SDS-guided refinement}
Next, we discuss the relation between the stage-2 alignment and SDS-guided refinement. While both of them can model the correlation between text and image, they have significant differences in their objectives and methods.

First, the stage-2 alignment aims to search for a text-consistent feature in the latent feature space of the SVR model, given the generated features, and then use the SVR model to produce a 3D shape. Thus, decoder D is frozen, which also helps to efficiently generate shapes that follow the text prompt. As shown in Figures 5(b), 13 (i), and 16, the stage-2 alignment can generate plausible shapes in only 85 seconds. Moreover, the results from stage 2 can serve as an initialization for stage 3 (e.g., SDS-guided refinement) for further improving the efficiency and generation quality of stage 3. Besides, the text conditional diffusion (i.e., diffusion prior) in stage 2 aims to produce diverse features from the text aligned with the corresponding CLIP image features, which then serve as the input to the SVR model for generating diverse 3D shapes consistent with the text. However, the stage-2 alignment is limited by the generation space of the pre-trained SVR model, so it cannot generate 3D shapes beyond the generative capability of the pre-trained SVR model. This is why stage 3 is needed, as explained below.

Second, our SDS-guided refinement in stage 3 employs a pre-trained text-to-image diffusion model to supervise the training of the text-to-shape generation model, which enables it to generate high-quality 3D shapes beyond the generative capability of the SVR model. In addition, the SDS-guided refinement aims to create novel structures and textures beyond the training dataset, and thus, it updates the decoder $D$ given the previous created coarse 3D shape from stage 2.
Besides, thanks to the supervision from a pre-trained diffusion model that allows updating the decoder D, the SDS-refinement stage can create out-of-distribution results beyond the training data, such as the police car in Figure 10 and the hamburger in Figure 12. In other words, the SDS-guided refinement optimizes a 3D shape by updating the network parameters of the SVR decoder instead of searching for a text-consistent feature in the latent space and reusing the SVR decoder.

\begin{figure}
\centering
\includegraphics[width=0.99\columnwidth]{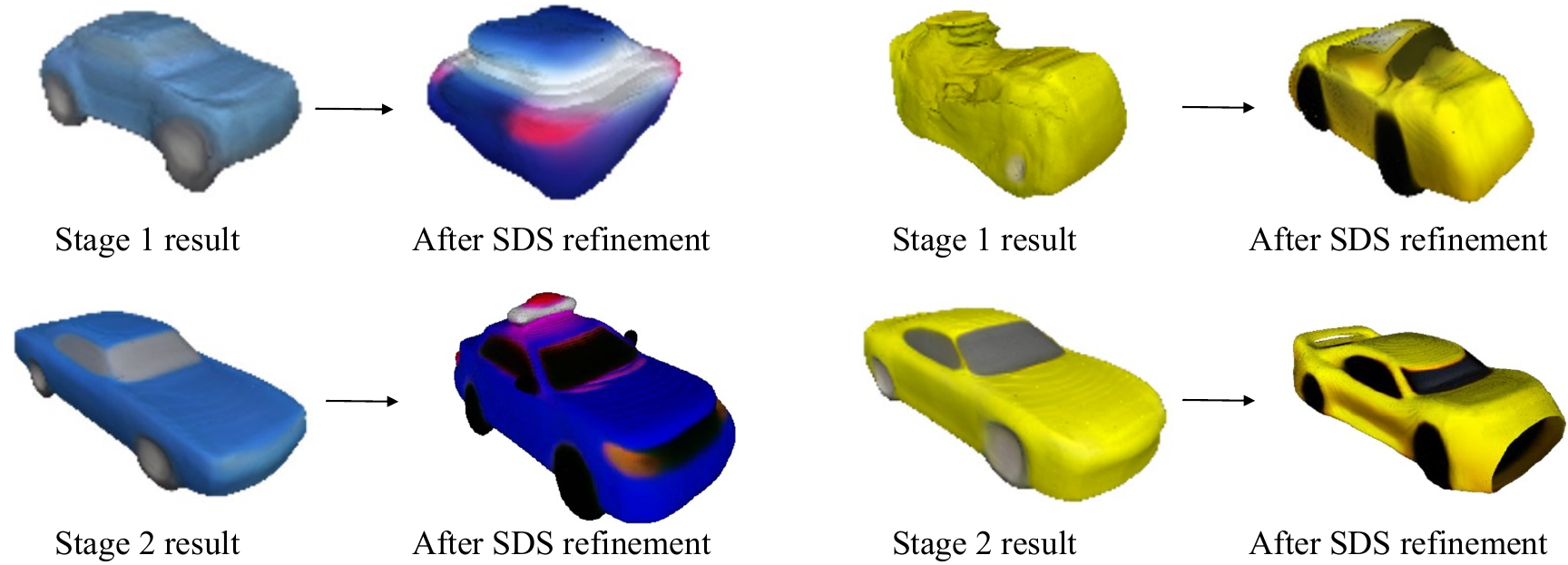}
\vspace*{-2mm}
\caption{
The result of the stage-1 alignment is not sufficient for the stage-3 SDS-guided refinement (top), whereas the stage-2 alignment can provide a good 3D prior to facilitate the SDS-guided refinement (bottom).
}
\label{fig:relation}
\vspace*{-2.5mm}
\end{figure}

Third, the relationship between them can be summarized as follows. The SDS-guided refinement can benefit from the stage-2 alignment, which can provide a better 3D prior. To demonstrate this point, we create a baseline that performs the SDS-guided refinement initialized with the stage-1 alignment results, instead of the stage-2 alignment results. As shown in Figure~\ref{fig:relation}, the stage-2 alignment result can provide a good 3D prior to help the SDS-guided refinement produce better results, while the stage-1 alignment result cannot.

\subsubsection{SDS-guided stylization}

Furthermore, this module also enables text-guided stylization to complement the CLIP-guided stylization presented in Section~\ref{sec:clip_stylization}. Specifically, given a 3D shape $S$ generated by our two-stage feature-space alignment and a text prompt $T$ that describes the target style, the SDS-guided  stylization procedure can incorporate the semantic attributes of $T$ into $S$, as illustrated in Figure~\ref{fig:overview_style} (b).

\noindent\textbf{Discussion on different stylization approaches}
\label{sec:discussion_on_stylization}
We presented three text-guided 3D shape stylization alternatives: texture stylization, shape-and-texture stylization, and SDS-guided stylization. Each has its own pros and cons. We compare the three stylization approaches using the same text prompts in Figure~\ref{fig:style}.


First, texture stylization mainly changes the texture style of the generated shape and preserves its own structure and functionality; see Figures~\ref{fig:style} (a) and~\ref{fig:style_compare} (b).
It can create more realistic textures compared with the other two stylization approaches; see``mahogany chair'' in Figure~\ref{fig:style_compare} (a). Also, it can handle abstract text descriptions (``sunset'') in Figure~\ref{fig:style_compare} (a), while shape-and-texture stylization creates unsatisfactory textures and SDS-guided stylization creates an empty shape. The simple case (``mahogany chair'') and the challenging case (``sunset sofa'') show that texture stylization 
surpasses the other two options when the user intends to paint the shape with plausible textures. However, texture stylization  
may result in shape-texture misalignment if the given shape and texture have misaligned structures (see Figure ~\ref{fig:style}: ``peach chair'').

Unlike the texture stylization, the other two approaches focus on creating novel and imaginary structures, giving rise to more plausible generative results (see Figure~\ref{fig:style} (b, c)). The CLIP-guided shape-and-texture stylization can produce better texture in some cases (“A rose chair”) while the SDS-guided stylization performs better in other cases (“Avocado boat”). Also, the SDS-guided stylization is capable of producing stylized 3D shapes that better capture the semantic concepts of the given style with better fidelity, see Figure~\ref{fig:style} (“An avocado chair” and “Tulip boat”). However, it may sacrifice the functionality of the generated shapes; see Figure~\ref{fig:style} (“An orchid chair”) while CLIP-guided shape-and-texture stylization can better balance the style concept and functionality (“An orchid chair”). 

In summary, there is a trade-off between preserving the functionality of 3D shapes and capturing the target style. To address this, we offer three options for users to choose from. Texture stylization is a good choice if the shape functionality is a top priority. Shape-and-texture stylization can encourage better consistency between texture and shape and is capable of generating novel structures. SDS-guided stylization can produce stylized 3D shapes with a higher fidelity according to the target style but at the expense of sacrificing their functionalities. {We hope our exploration will inspire more research efforts in the future for simultaneously achieving functionality preservation and style creation.} 

\subsection{Compatibility with Different SVR Models}
\label{sec:compatible} 

In addition to DVR, our two-stage feature-space alignment can work with a variety of SVR models. For instance, it can be easily integrated with advanced methods, SS3D~\cite{alwala2022pre} and GET3D~\cite{gao2022get3d}, two recent generative models for 3D shape generation. 
{SS3D is capable of generating 3D shapes for a wide range of categories and GET3D can generate striking 3D shapes in superior quality. }
By replacing  $E_{\text{S}}$ and $D$ in Figure~\ref{fig:overview} with the encoder and decoder of SS3D or GET3D, our model can  be integrated with them 
 and produce shapes of  more categories or higher qualities. 
During training, we can adopt a similar pipeline depicted in Figure~\ref{fig:overview} to enable text-to-shape generation. 
For SS3D, in stage-1 training, we use their training objectives to replace $\mathcal{L}_\text{D}$ (see Section~\ref{sec:two-sgage}), which uses single-view in-the-wild  images beyond the ShapeNet categories without their poses.
For GET3D, we first generate paired image-shape data by rendering images from its generated 3D shapes for training our two-stage feature-space alignment pipeline. 
In a nutshell, our approach is scalable and compatible with various SVR models and can potentially benefit from other new approaches in the future.

\section{Experiments}
\label{sec:results}


\subsection{Dataset}
\label{sec:implementation}


To train our DreamStone framework, we use both synthetic and real-world datasets, ShapeNet~\cite{shapenet2015} (13 categories) and CO3D~\cite{reizenstein2021common} (50 categories), respectively. 
We further extend the generative capability beyond the above categories by adopting SS3D~\cite{alwala2022pre} and fine-tuning our model using SDS. 
For conducting quantitative and qualitative evaluations, we create a test set with four pieces of texts per category in the ShapeNet dataset. 


\subsection{Implementation Details}
To train the two-stage feature-space alignment model, we first train the stage-1 mapping with the learning rate of $1e^{-4}$ for 400 epochs. Then at test time, we further train the stage-2 alignment for 20 iterations. On average, this process takes around 85 seconds using one GeForce RTX 3090 Ti GPU. Optionally, we further refine $S$ with SDS loss for about $40$ epochs or text-guided stylization for about $30-50$ epochs. 
Our hyperparameters, including $\lambda_{M}$, t, $\lambda_{bg}$, $m$, and $\tau$, are set empirically to 0.5,  0.5, 10, 10, and 0.5, respectively, based on a small validation set. 

\subsection{Metrics}
\subsubsection{Metric for shape generation quality}

For quantitative evaluation, we adopt the Fréchet Inception Distance (FID)~\cite{heusel2017gans} between a set of five rendered images from different camera views for each shape and a set of ground-truth images from ShapeNet.
We use the official model with InceptionNet pre-trained on ImageNet for FID evaluation, as it is a widely adopted metric for evaluating the realism and quality of generative models. We do not train an FID model on ShapeNet, as the size of the dataset is too small to train an effective FID model like that trained on ImageNet.
Additionally, we randomly sample 2,600 images from the ShapeNet dataset as ground-truth images for FID evaluation, rather than using images from ImageNet, to more accurately evaluate the similarity between the generated shapes and the ShapeNet ground truths.

\begin{figure}
\centering
\includegraphics[width=0.99\columnwidth]{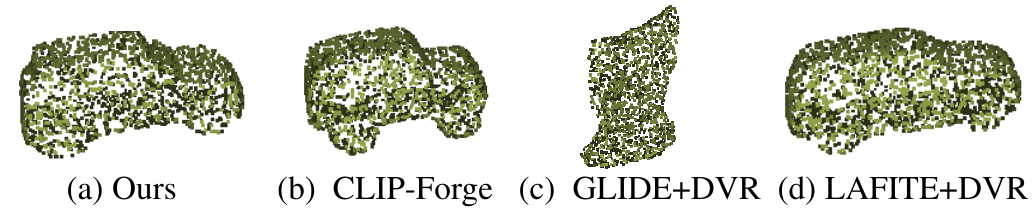}
\vspace*{-1.5mm}
\caption{Visualize point clouds of different methods for FPD evaluation.}
\label{fig:point_cloud}
\vspace*{-2.5mm}
\end{figure}

Besides adopting FID, we also utilize the metric Fréchet Point Distance (FPD) proposed in~\cite{liu2022towards} to measure the shape generation quality without texture. To evaluate FPD, We first extract 3D point clouds from the generated shapes without color (see Figure~\ref{fig:point_cloud}) and then evaluate them. It is worth mentioning that Dream Fields~\cite{jain2021zero} does not generate 3D shapes directly, so we could not evaluate it using FPD in this aspect.

\vspace*{-3pt}
\subsubsection{Human perceptual evaluation setup}

Further, we conduct a human perceptual evaluation to assess the consistency between the generated shapes and the input text. 
To begin with, we collect the generated results. 
For each input text, we create 14 results in total from the four existing works, eight baseline methods, and our predecessor work ISS~\cite{liu2023iss} and our DreamStone; see Section~\ref{sec:existing} and Section~\ref{sec:ablation} for details of each approach.
Then, we invite 10 volunteers with normal vision to participate in the evaluation, including 3 females and 7 males whose ages are in the range of 19 to 58. The generated results are shown to the participants in random order without any hint on how they are created. Then the volunteers are asked to give a score to indicate whether the candidate shape matches the input text, where 1 means a perfect match, 0.5 means a partial match, and 0 indicates a poor match.  
At last, we sum up the total score $s$ for each approach from all participants and calculate $s/n$ as the metric ``Consistency Score'', where $n=10$ means the number of collected samples. 

\subsection{Comparisons with Existing Works}\label{sec:existing}

Next, we perform qualitative and quantitative comparisons of four state-of-the-art works~\cite{sanghi2021clip,jain2021zero,mohammad2022clip,poole2022dreamfusion}, our predecessor work ISS~\cite{liu2023iss}, and our DreamStone. For DreamFusion~\cite{poole2022dreamfusion}, as there are no official codes available, we use the latest version of a third-party implementation called Stable-DreamFusion~\cite{stable-dreamfusion}. For the other works, we use their official codes on GitHub to generate shapes.

\subsubsection{Quantitative comparisons} 
According to the quantitative comparisons presented in Table~\ref{tab:all}, our result ``DreamStone'' outperforms all the existing works by a considerable margin in terms of all the evaluation metrics, as shown in Table~\ref{tab:all}. Specifically, the superior performance on FID and FPD demonstrates that our generative results have better quality in terms of the texture and 3D topology. In addition, the higher Consistency Score indicates that DreamStone can generate shapes with better consistency with the input text. The results of ``A/B/C Test'' and ``A/B Test'' will be discussed later in Section~\ref{sec:abtest}. 

\subsubsection{Qualitative comparisons}

\begin{figure*}
\centering
\includegraphics[width=0.99\textwidth]{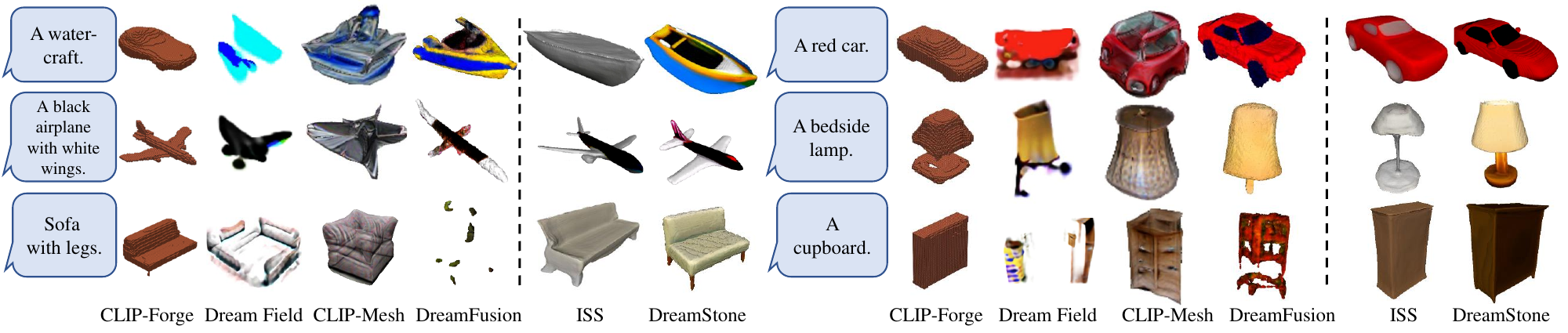}
\vspace*{-1.5mm}
\caption{Qualitative comparisons with existing works. Note that DreamStone is built upon ISS~\cite{liu2023iss} and incorporates DP and SDS. 
}
\label{fig:sota}
\vspace*{-2.5mm}
\end{figure*}

\begin{table*}
\centering
\caption{Comparisons with existing works and our baselines. $^*$: We use Stable-Dreamfusion~\cite{stable-dreamfusion} for implementation.
}
\scalebox{0.95}{
  \begin{tabular}{cccccccc}
    \toprule
  \multirow{2}{*}{\rotatebox{0}{
Method Type
}} & \multirow{2}{*}{\rotatebox{0}{
Method
}} & \multirow{2}{*}{\rotatebox{0}{
FID ($\downarrow$)
}} & \multirow{2}{*}{\rotatebox{0}{
Consistency Score (\%) ($\uparrow$) 
}}& \multirow{2}{*}{\rotatebox{0}{
FPD ($\downarrow$) 
}} & A/B/C Test & A/B Test  \\ 
&&&&& (for two-stage alignment) & (for DreamStone) \\
    \midrule
    \multirow{4}{*}{\rotatebox{0}{
Existing works
}}&  CLIP-Forge~\cite{sanghi2021clip}  & 162.87 
&  41.83 $\pm$ 17.62 & 37.43 & 8.90 $\pm$ 4.12 & N.A. \\
    & Dream Fields~\cite{jain2021zero} & 181.25 &  25.38 $\pm$ 12.33  & N.A. & N.A. & N.A. \\
    &CLIP-Mesh~\cite{mohammad2022clip} & 188.09 & 40.27 $\pm$ 8.82 & 40.27 & N.A. & N.A. \\
    &DreamFusion~\cite{poole2022dreamfusion}$^*$ & 159.04 & 38.36 $\pm$ 9.12 & 36.44 & N.A. & 7.00 $\pm$ 2.64  \\
    \midrule
    \multirow{6}{*}{\rotatebox{0}{
Ablation studies
}}  & $E_{\text{I}}$+$D$ & 181.88 & 20.97 $\pm$ 13.59 & 38.61 & N.A. & N.A.\\
    & w/o stage 1 & 222.96 & 1.92 $\pm$ 2.22  & 79.41 &  N.A. & N.A. \\
    & w/o stage 2 & 202.33 & 29.52 $\pm$ 14.86  & 41.71 &  N.A.& N.A.\\
    & w/o $\mathcal{L}_{\text{bg\_1}}$ & 149.45 & 29.45 $\pm$ 14.67 & 40.85 & N.A.& N.A.\\
    & w/o $\mathcal{L}_{\text{bg\_2}}$ & 156.52  & 31.55 $\pm$ 8.87 & 38.31 & N.A.& N.A. \\
    & w/o $\mathcal{L}_{\text{bg}}$ & 178.34 & 30.96 $\pm$ 15.49 & 40.98 & N.A. & N.A.\\
    \midrule
    \multirow{2}{*}{\rotatebox{0}{
Text2Image+SVR
}}  & GLIDE~\cite{nichol2021glide}+DVR~\cite{niemeyer2020differentiable} & 212.41 & 8.85 $\pm$ 7.94 & 41.33 & N.A. & N.A.\\
    & LAFITE~\cite{zhou2021lafite}+DVR~\cite{niemeyer2020differentiable} & 135.01 & 52.12 $\pm$ 11.05  & 37.55 & 11.70 $\pm$ 4.11 & N.A.\\
    \midrule
    Our earlier work & ISS~\cite{liu2023iss} & 124.42 $\pm$ 5.11 & 60.0 $\pm$ 10.94  & 35.67 $\pm$ 1.09 & \textbf{21.70 $\pm$ 5.19} & N.A. \\
    \midrule
    Ours & DreamStone & \textbf{114.34} & \textbf{70.77 $\pm$ 8.38} &  \textbf{30.92}   &  N.A.& \textbf{31.80 $\pm$ 7.53} \\
    \bottomrule
  \end{tabular}
  }
  \label{tab:all}
\end{table*}

\noindent\textbf{Comparison with state of the arts.} Then, we compare the generative results of our DreamStone with four existing works and our predecessor work ISS~\cite{liu2023iss}. The qualitative comparisons  are shown in Figure~\ref{fig:sota}. We observe that CLIP-Forge~\cite{sanghi2021clip} can only produce low-resolution shapes without color and texture, and some of its generated shapes do not well align with the input text, for instance, ``a watercraft''. 
Dream Fields~\cite{jain2021zero} fails to generate desired shapes in most evaluated cases. 
Also, CLIP-Mesh~\cite{mohammad2022clip} is unable to generate fine-grained topology in some cases such as ``a black airplane with white wings''. 
Besides, Stable-Dreamfusion~\cite{stable-dreamfusion} has inferior performance in terms of surface quality (``a black airplane with white wings''), topology faithfulness (``a cupboard''), and generative efficiency.  
Despite that our predecessor work ISS~\cite{liu2023iss} can produce 3D shapes with better topology faithfulness and less time-consuming, the details of the results are still far from satisfactory, e.g., the rearview mirror on ``a red car''. In contrast, our DreamStone outperforms all the existing works by a large margin in terms of generative quality, consistency with the input text, and details on the generated shape, as shown in Figure~\ref{fig:sota}. 





\begin{figure*}
\centering
\includegraphics[width=0.99\textwidth]{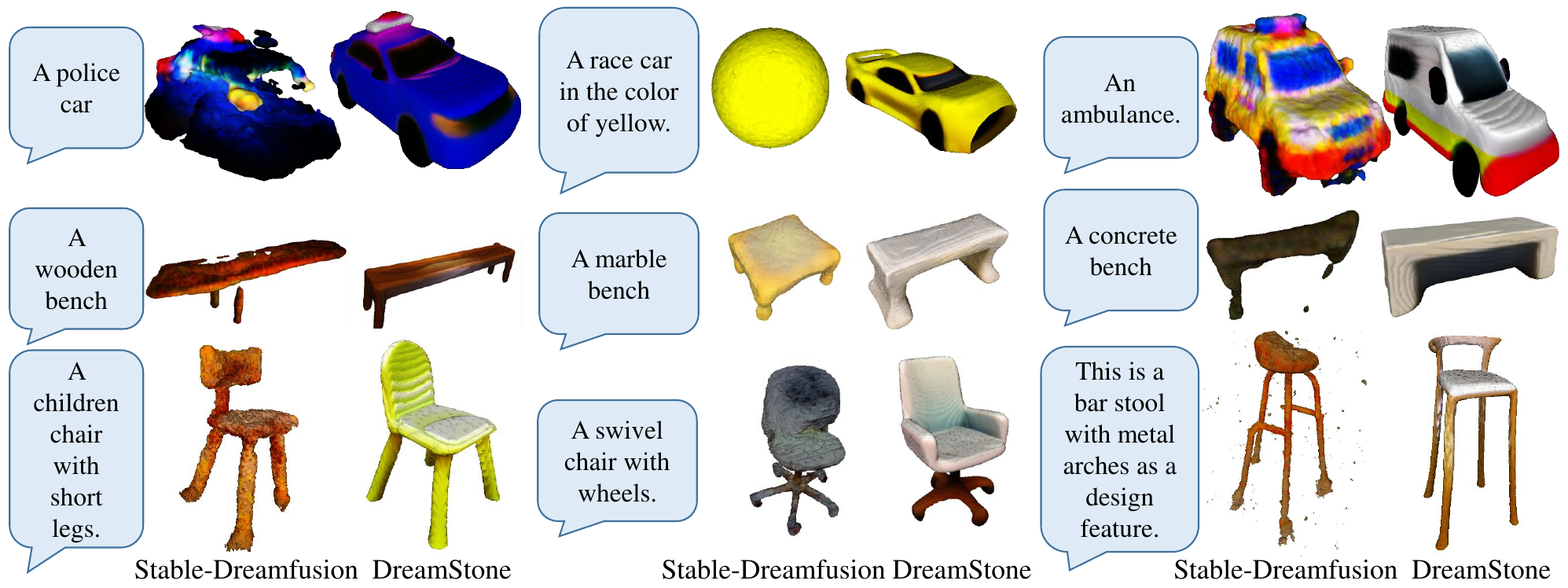}
\vspace*{-2.5mm}
\caption{Results of Stable-Dreamfusion and our DreamStone. Note that DreamStone is built upon ISS~\cite{liu2023iss} and incorporates DP and SDS.}  
\label{fig:in-category}
\vspace*{-3.5mm}
\end{figure*}
\begin{figure*}
\centering
\includegraphics[width=0.99\textwidth]{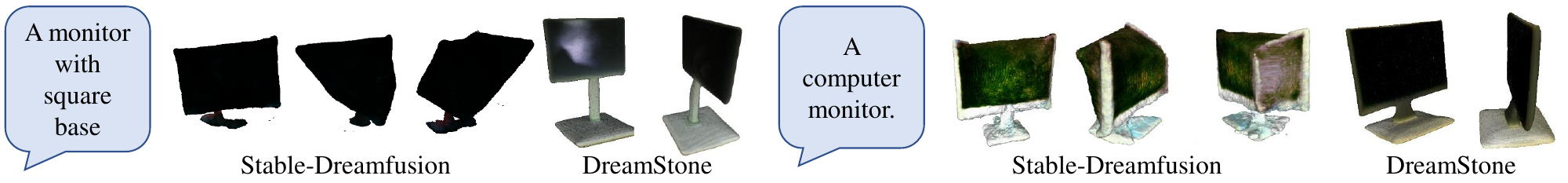}
\vspace*{-2.5mm}
\caption{Stable-DreamFusion suffers from the ``multi-face Janus problem'', and our DreamStone mitigates this issue by leveraging the 3D prior. } 
\label{fig:multiface}
\vspace*{-3.5mm}
\end{figure*}

\vspace{0.1in}\noindent\textbf{Comparison with DreamFusion.} To provide a further comparison with the most recent work DreamFusion~\cite{poole2022dreamfusion}, we show additional generative results from Stable-DreamFusion~\cite{stable-dreamfusion} and our DreamStone in Figure~\ref{fig:in-category}. Unlike Stable-DreamFusion, which optimizes the shape directly using SDS without a 3D prior, our DreamStone utilizes the 3D prior learned by our two-stage feature-space alignment, improving the generative performance in terms of avoiding failure modes (e.g., ``a race car in the color of yellow''), enhancing the surface quality (e.g., ``an ambulance''), and improving the 3D topology faithfulness (e.g., ``a swivel chair with wheels'').
In addition, our DreamStone mitigates the ``multi-face Janus problem'' in Stable-DreamFusion, where the generated shapes, e.g., the monitors in Figure~\ref{fig:multiface}, can have multiple frontal views when viewed from different viewpoints. On the contrary, our DreamStone is able to generate faithful 3D shapes leveraging the 3D prior learned in our two-stage feature-space alignment, see Figure~\ref{fig:multiface} ``DreamStone''
Moreover, our DreamStone can significantly reduce the generation time compared to DreamFusion~\cite{poole2022dreamfusion}, which needs more than an hour to create a single 3D shape. As shown in Figure~\ref{fig:in-category}, our DreamStone can generate a 3D shape in only 22.5 minutes on average (30 training epochs, each 45 seconds on a single RTX-3090 GPU). Together with the two-stage feature-space alignment that provides the 3D prior, our DreamStone still takes less than 25 minutes in total, which is much faster than DreamFusion, which takes 1.5 hours on average.

\begin{figure*}
\centering
\includegraphics[width=0.99\textwidth]{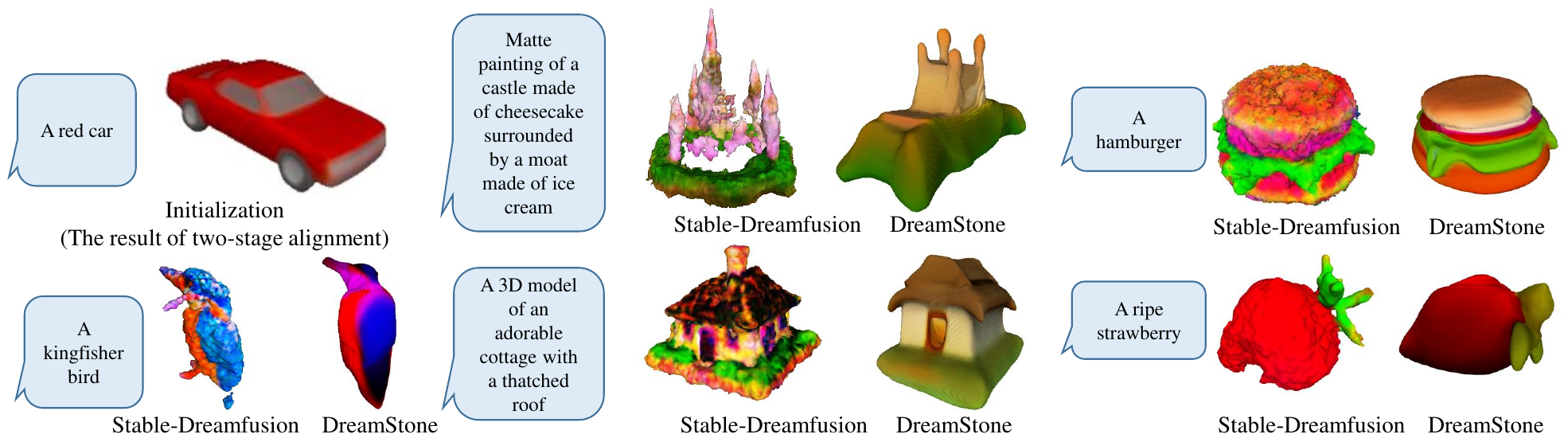}
\vspace*{-2.5mm}
\caption{With a randomly selected shape as initialization (``a red car''), DreamStone can generate a wide range of 3D shapes beyond the training categories. Note that DreamStone is built upon ISS~\cite{liu2023iss} and incorporates DP and SDS. }
\label{fig:out-category}
\vspace*{-3.5mm}
\end{figure*}

\vspace{0.1in}\noindent\textbf{Generalization ability to novel categories.} Another notable advantage of our DreamStone is its ability to generate 3D shapes in novel categories beyond the training data. As depicted in Figure~\ref{fig:out-category}, starting from a randomly chosen shape ``a red car'' from our two-stage feature-space alignment, DreamStone is capable of deforming it into various 3D shapes (Figure~\ref{fig:out-category}) in a broad range of categories. 
It is worth noting that the quality of generated shapes can benefit from 3D priors of unrelated categories. 
For instance, a ``bird'' can be generated from using  a ``car'' as prior. 
This might be caused by the smoothness priors enforced by the initialization model, which is further used by the subsequent SDS process to produce high-quality surface. 
This demonstrates the generalization ability of our method in generating diverse and plausible novel 3D shapes, even for input texts beyond the training categories.

\subsection{Ablation Studies}\label{sec:ablation}

\subsubsection{Baseline setups}\label{sec:ablation:setup}

In addition, we develop several baselines to evaluate the effectiveness of different components in our model. 
%
\begin{itemize}[leftmargin=0.5cm]
\item $E_{\text{I}}+D$: This is the baseline, where we get the CLIP image feature $f_{\text{I}}$ using $E_{\text{I}}$, and optimize $D$ to generate 3D shapes from $f_{\text{I}}$ without using the two-stage feature-space alignment.
\item w/o (without) stage 1: 
we ablate the stage-1 alignment and optimize the stage-2 alignment with a randomly initialized $M$.
\item w/o stage 2: we directly generate the shape with the mapper $M$ after stage 1, without performing the stage-2 optimization.
\item w/o $\mathcal{L}_{\text{bg\_1}}$: 
removing $L_\text{bg}$ in stage-1 alignment.
\item w/o $\mathcal{L}_\text{bg\_2}$: 
removing $L_{\text{bg}}$ in stage-2 alignment.
\item w/o $\mathcal{L}_\text{bg}$: 
removing $\mathcal{L}_{\text{bg}}$ in both stages.
\item GLIDE+DVR: 
using a recent zero-shot text-to-image generation method GLIDE~\cite{nichol2021glide} to first generate image $I$ from $T$, and then using DVR~\cite{niemeyer2020differentiable} to generate $S$ from $I$.
\item LAFITE+DVR: 
we train a recent text-guided image generation approach LAFITE~\cite{zhou2021lafite} on ShapeNet dataset, produce an image $I$ from $T$, and then generate $S$ from $I$ using DVR~\cite{niemeyer2020differentiable}.
\end{itemize}

The first six baselines are designed to evaluate the effectiveness of  modules in our framework and the last two baselines utilize advanced text-guided 2D image generation methods to first generate images and then use an SVR model to generate shapes. 
Note that we still adopt DVR as the SVR model for fair comparisons. 

\subsubsection{Quantitative and qualitative comparisons }



\begin{figure*}
\centering
\includegraphics[width=0.99\textwidth]{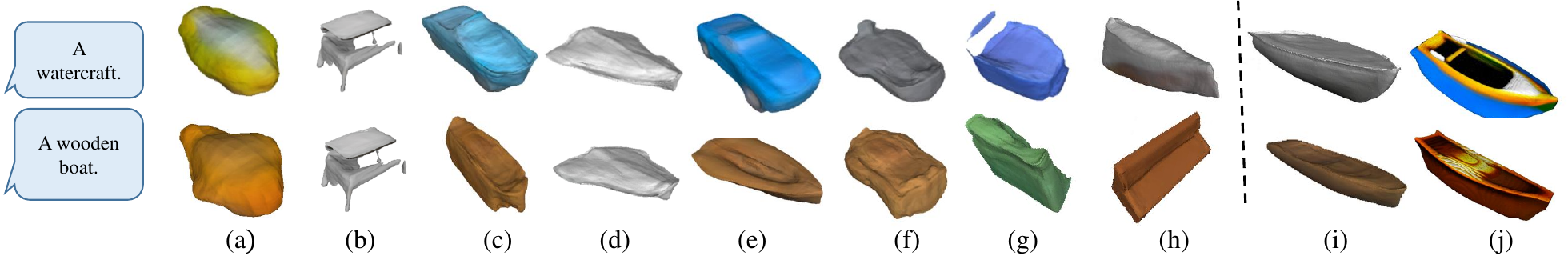}
\vspace*{-1.5mm}
\caption{Ablation studies. (a) $E_\text{I} + D$; (b) w/o stage-1; (c) w/o stage-2; (d) w/o $L_{bg\_1}$; (e) w/o $L_{bg\_1}$; (f) w/o $L_{bg}$; (g) GLIDE+DVR; (h) Lafite + DVR; and (i) baseline, which indicates ISS~\cite{liu2023iss}; and (j) baseline~\cite{liu2023iss} + SDS.
}
\label{fig:quality}
\vspace*{-2.5mm}
\end{figure*}

\begin{figure*}
\centering
\includegraphics[width=0.99\textwidth]{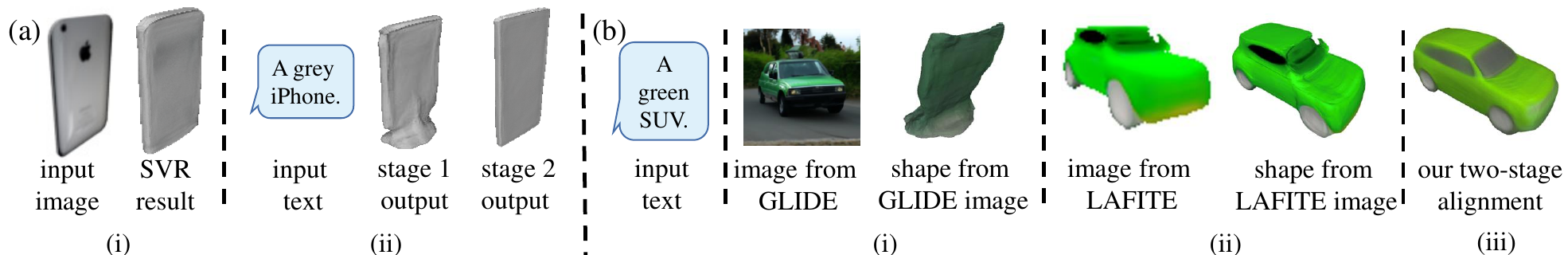}
\vspace*{-1.5mm}
\caption{A further study of baselines ``w/o stage 2'', ``GLIDE+DVR'', and ``LAFITE+DVR''.
%
(a) ``w/o stage 2'' generates a reasonable shape (``SVR result'') from the input image, but produces a low-quality shape (``stage 1 output'') when using the text as input; further fine-tuning our model with stage-2 alignment allows us to generate a more plausible shape using the text (``stage 2 output'').
(b) GLIDE and LAFITE tend to generate out-of-domain and low-quality images, respectively, leading to inferior performance in the subsequent image-based 3D generation.
%
%
%
%
}
\label{fig:ablation_discuss}
\vspace*{-2.5mm}
\end{figure*}

The qualitative results of baseline methods are shown in Figure~\ref{fig:quality}. We summarize our key observations as below:



\begin{itemize}[leftmargin=0.5cm]
\item $E_{\text{I}}+D$: As seen in column (a) of Figure~\ref{fig:quality}, the generated results from CLIP space $\Omega_{\text{I}}$ have inferior texture and shape structure fidelity due to the inferior ability of $E_\text{I}$ in capturing image details.

\item w/o stage 1: Figure~\ref{fig:quality} (b) shows that the produced shapes are almost the same for any given text without adopting stage-1 alignment. This happens because $M$ maps text feature $f_\text{T}$ to nearly the same feature even with stage-2 alignment enabled. This demonstrates the necessity of stage-1 alignment to provide good initialization for stage-2 test-time optimization.

\item w/o stage 2: Figure~\ref{fig:quality} (c) indicates that the model may fail to align $f_S$ and $f_{\text{T}}$ well without stage 2. This can be further illustrated in Figure~\ref{fig:ablation_discuss} (a). Without using stage 2,  the model fails to generate a reasonable shape with text as input but successes in generating 3D shapes from a single image. 
After applying stage 2, a plausible phone can be produced using the text (see ``stage 2 output'').
%

\item w/o $\mathcal{L}_{\text{bg\_1}}$, w/o $\mathcal{L}_{\text{bg\_2}}$, w/o $\mathcal{L}_{\text{bg}}$: Columns (d, e, f) of Figure~\ref{fig:quality} show that stage-2 alignment cannot work properly without $\mathcal{L}_{\text{bg}}$ in either stage-1 or stage-2 alignment or both due to the lack of foreground awareness. Even though stage-1 alignment has already encouraged the background to be white, we still need this loss in stage 2 to obtain satisfying results.

\item GLIDE+DVR: The performance of GLIDE+DVR (see Figure~\ref{fig:ablation_discuss} (b)) is poor because of the large domain gap between the training data of DVR and the images generated by GLIDE~\cite{nichol2021glide}.
%

\item LAFITE+DVR: In Figure~\ref{fig:quality} (h), some shapes produced by this baseline  do not match the given texts because of the semantic gap between $f_{\text{I}}$ and $f_{\text{T}}$ (e.g., ``a wooden boat''). Also, the appearance can be coarse (Figure~\ref{fig:ablation_discuss} (b)) because of the error accumulation of the isolated two steps, i.e., LAFITE (Figure~\ref{fig:ablation_discuss} (b) ``image from LAFITE'') and DVR (Figure~\ref{fig:ablation_discuss} (b) ``shape from LAFITE image''). Despite these shortcomings, generating images and shapes in a subsequent manner remains a strong baseline that is a valuable direction for future research. 

\item Two-stage alignment: 
Column (i) of Figure~\ref{fig:quality} shows that our two-stage feature space alignment can generate plausible shapes and textures consistent with text descriptions, beyond all the above baselines. However, the generative details are still not very satisfying. 
\item Ours (DreamStone): Column (j) of Figure~\ref{fig:quality} demonstrates the superior capability of DreamStone to generate shapes and textures with a remarkable level of detail, outperforming all the baselines by a substantial margin.
\end{itemize}


\vspace*{-3pt}
\subsubsection{A/B/C test and A/B test}
\label{sec:abtest} 
We conduct an A/B/C test and an A/B test with 10 volunteers. For fair comparisons, the A/B/C test is designed to evaluate the approaches 
without SDS refinement, {\ie}, our two-stage feature-space alignment and two baselines that have the highest performance: CLIP-Forge~\cite{sanghi2021clip} and ``LAFITE+DVR''. Also, the A/B test aims to compare the approaches trained with SDS, including our DreamStone with DreamFusion~\cite{poole2022dreamfusion}.  
In this test, the results of the three approaches (per input text, a total of 52 texts) were displayed in a random order, and the participants were asked to choose their favorite one. 

The results of the A/B/C test, shown in Table~\ref{tab:all} ``A/B/C Test'', demonstrate that our two-stage feature-space alignment is the most preferred approach, outperforming CLIP-Forge by 143.8\% (computed as $(21.70-8.90)/8.90$) and ``LAFITE+DVR'' by 85.5\% (computed as $(21.70-11.70)/11.70$). In addition, the result of ``A/B test'' in Table~\ref{tab:all} shows that our DreamStone outperforms Stable-Dreamfusion by 354.3\% (computed as $(31.80-7.00)/7.00$) in terms of user preference. 




\vspace*{-5pt}

\subsection{More Analysis of Two-Stage Alignment}\label{sec:show}

Next, we evaluate the novelty and diversity of generated shapes, as well as the scalability of the proposed two-stage feature-space alignment.  





\begin{figure*}
\centering
\includegraphics[width=0.99\textwidth]{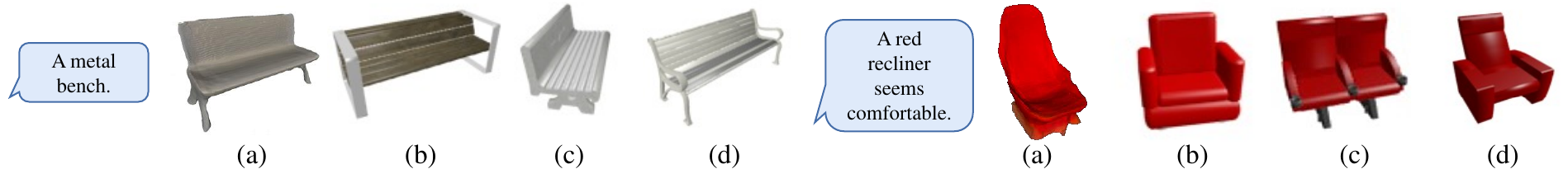}
\vspace*{-3.5mm}
\caption{Our two-stage feature-space alignment can create novel shapes that are not in the training set.
(a) displays our results and (b,c,d) are the top three closest shapes retrieved from the training set. Note that these 
shapes are generated with the two-stage feature-space alignment without the diffusion prior and SDS-guided refinement.}
\label{fig:novelty}
\vspace*{-3.5mm}
\end{figure*}

\begin{figure*}
\centering
\includegraphics[width=0.99\textwidth]{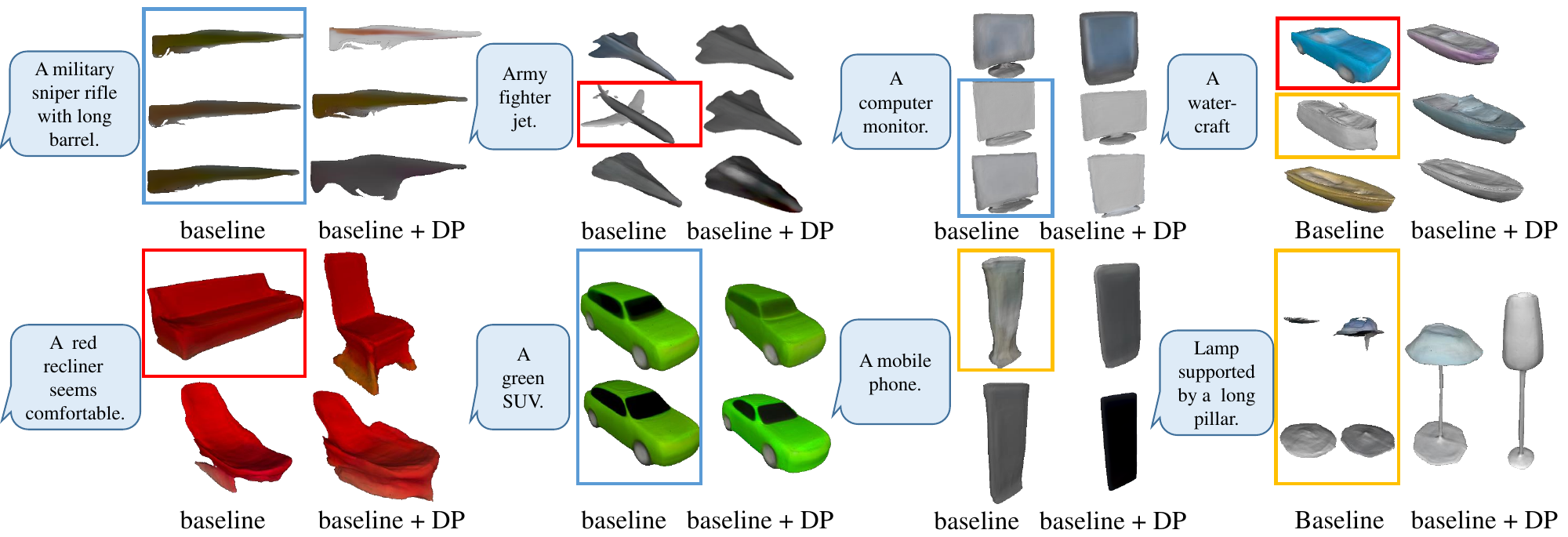}
\vspace*{-3.5mm}
\caption{When equipped with the diffusion prior, our two-stage alignment produces more diversified results with better text-shape consistency given the same set of input texts. 
Baseline: ISS~\cite{liu2023iss}. The color of the bounding boxes indicate the quality of the results.
Red indicates inconsistent with the input text.
Orange indicates low quality.
Blue indicates similar results that lack diversity.
Note that DP (diffusion prior) refers to the model that generates an associated CLIP image feature from the CLIP
text feature described in~\cite{ramesh2022hierarchical}.}
\label{fig:diversified}
\vspace*{-3.5mm}
\end{figure*}

\begin{figure*}
\centering
\includegraphics[width=0.99\textwidth]{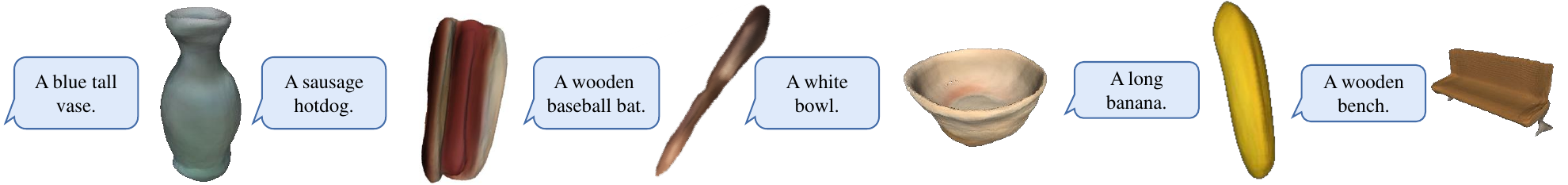}
\vspace*{-3.5mm}
\caption{Results of our two-stage feature-space alignment on the CO3D dataset.
These shapes are generated with the two-stage feature-space alignment without the diffusion prior and SDS-guided refinement.
} 
\label{fig:co3d}
\vspace*{-3.5mm}
\end{figure*}

\begin{figure*}
\centering
\includegraphics[width=0.99\textwidth]{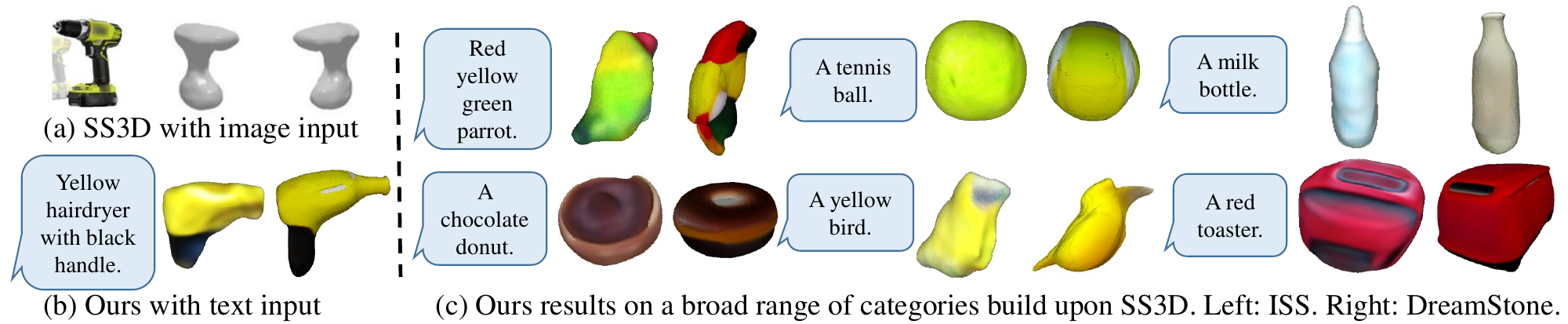}
\vspace*{-2.5mm}
\caption{After training on single images (without camera poses), our approach can generate shapes for a broad range of categories, by adopting~\cite{alwala2022pre}. 
The shapes on the left of (b) and (c) are the results of ISS~\cite{liu2023iss} generated with the two-stage feature-space alignment without the diffusion prior and SDS-guided refinement, and those on the right of (b) and (c) are generated by our DreamStone, which indicates ISS~\cite{liu2023iss} + DP + SDS. }
\label{fig:ss3d}
\vspace*{-3.5mm}
\end{figure*}

\vspace*{-3.5mm}


\begin{figure*}
\centering
\includegraphics[width=0.99\textwidth]{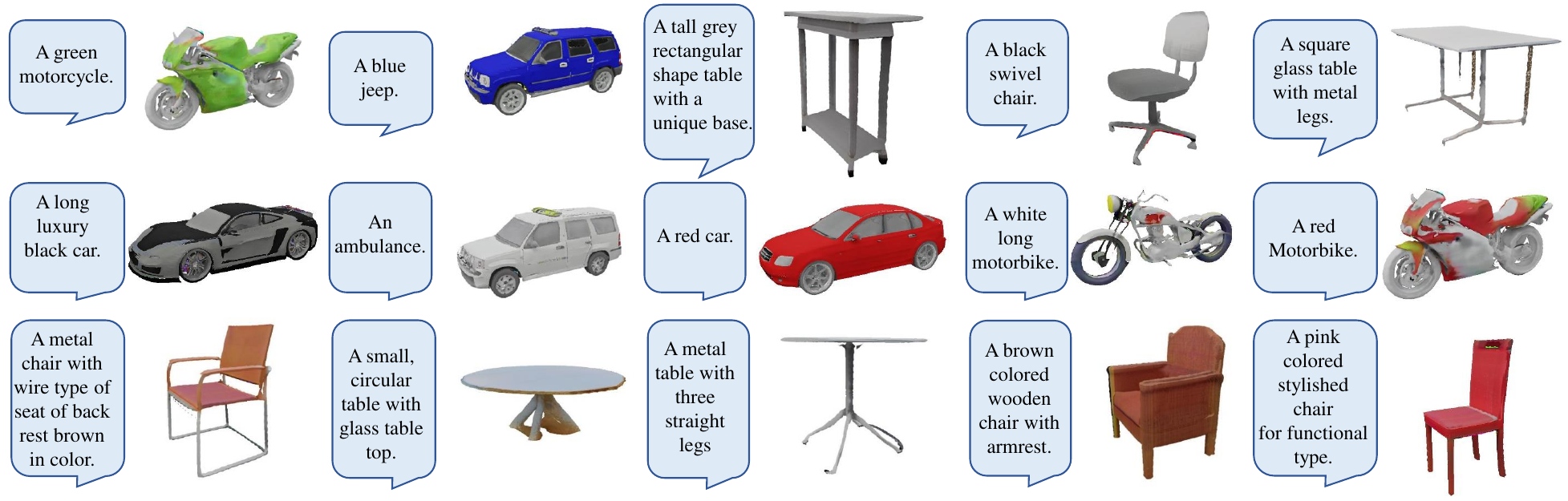}
\vspace*{-3.5mm}
\caption{
Our two-stage feature-space alignment can inherit the superior generative capability of GET3D~\cite{gao2022get3d} to generate high-quality 3D shapes even without the diffusion prior and SDS-guided refinement. }
\label{fig:get3d}
\vspace*{-3.5mm}
\end{figure*}

\begin{figure*}
\centering
\includegraphics[width=0.99\textwidth]{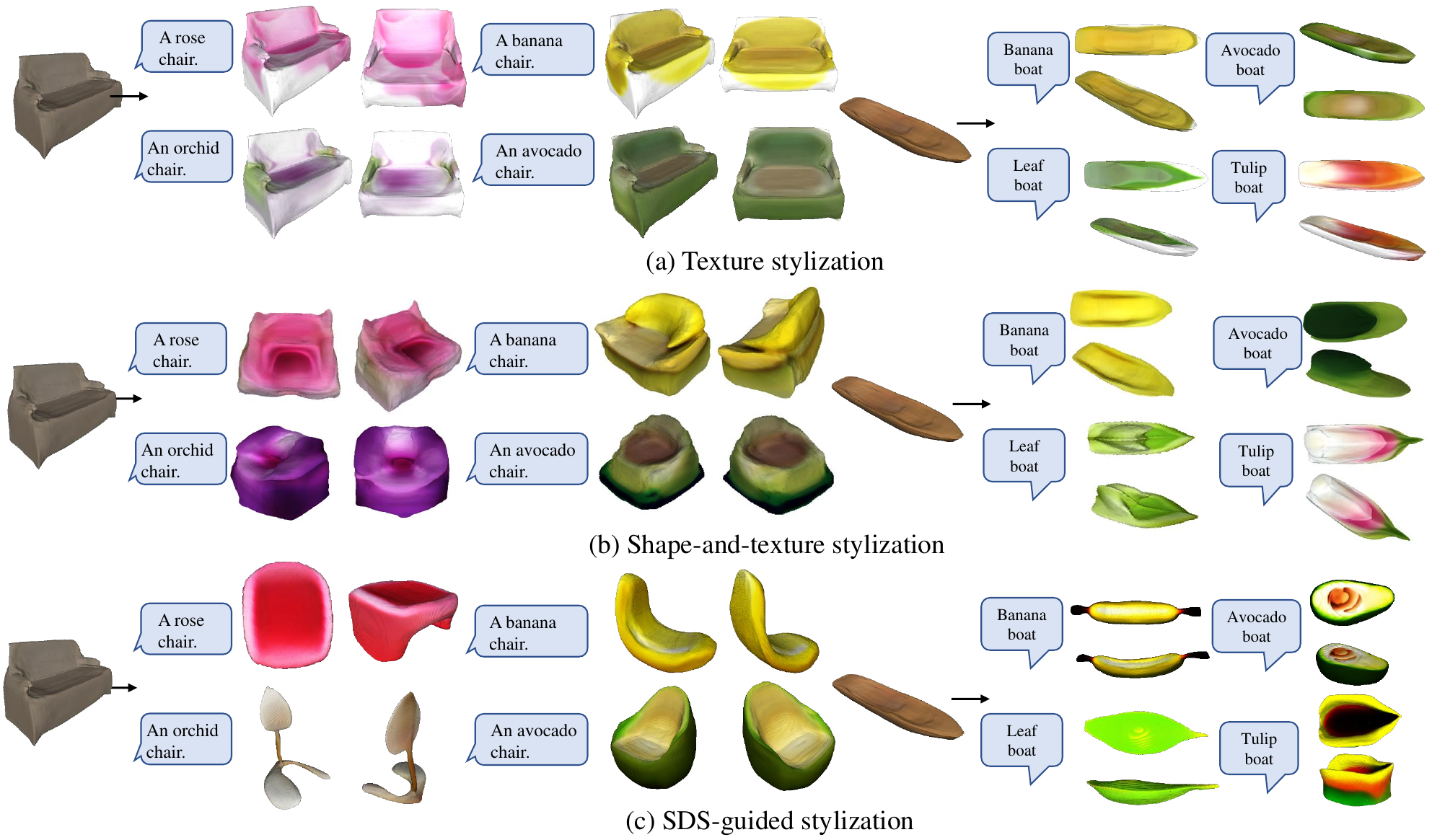}
\vspace*{-4mm}
\caption{Results of our text-guided stylization approach. (a) Texture stylization. (b) Shape-and-texture stylization. (c) SDS-guided stylization. }
\label{fig:style}
\vspace*{-2.5mm}
\end{figure*}

\vspace*{-2.5mm}
\begin{figure*}
\centering
\includegraphics[width=0.99\textwidth]{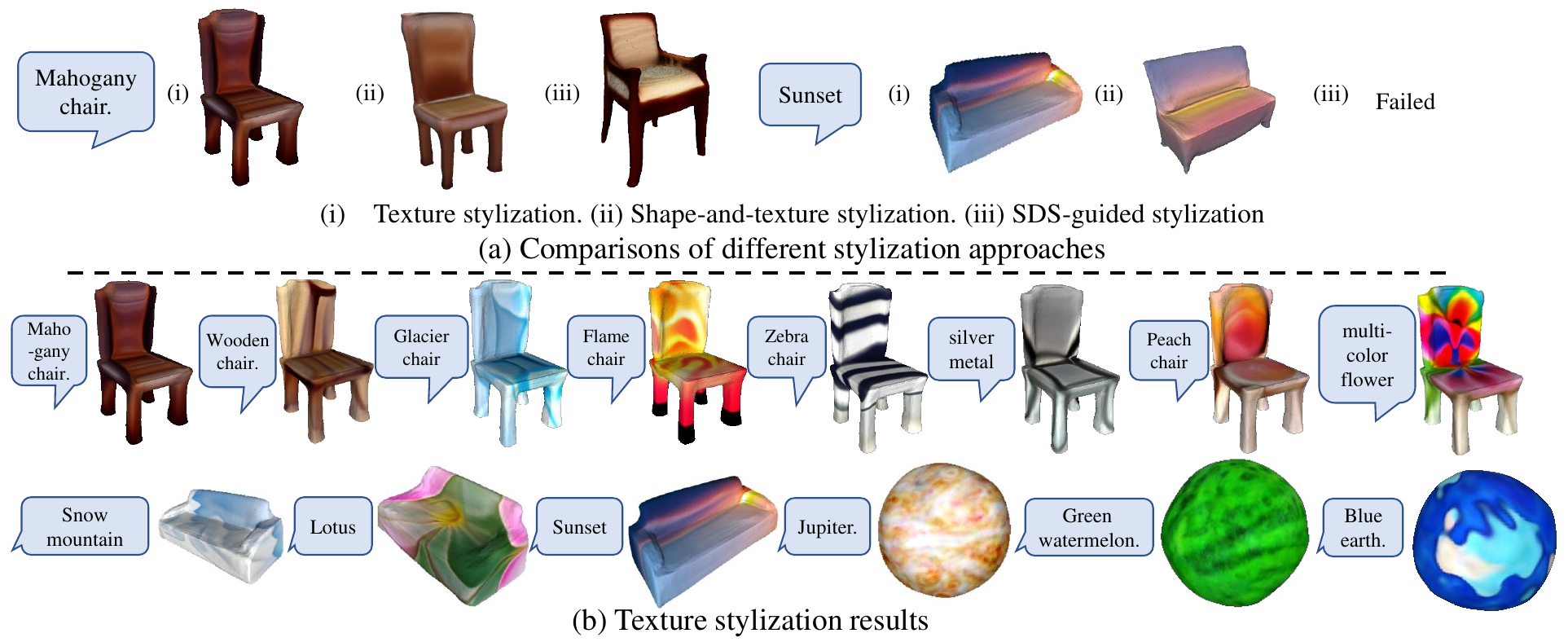}
\vspace*{-2.5mm}
\caption{
Results of texture stylization. (a) Texture stylization outperforms the other stylization approaches in terms of texture generation (mahogany chair) and handling abstract text descriptions (sunset). (b) Gallery of texture stylization. 
}
\label{fig:style_compare}
\vspace*{-2.5mm}
\end{figure*}

\begin{figure*}
\centering
\includegraphics[width=0.99\textwidth]{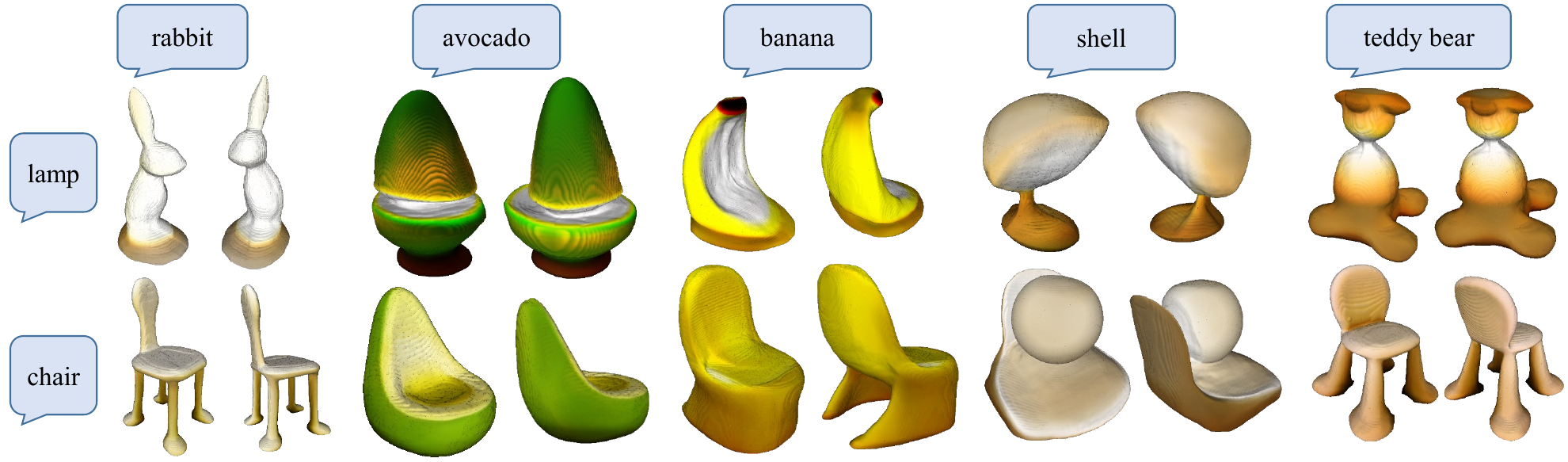}
\vspace*{-4mm}
\caption{Additional results of SDS-guided stylization. Two different views are rendered. The text prompt is ``A [shape] simulating a [style].'' } 
\label{fig:sds_style}
\vspace*{-3.5mm}
\end{figure*}

\begin{figure*}
\centering
\includegraphics[width=0.99\textwidth]{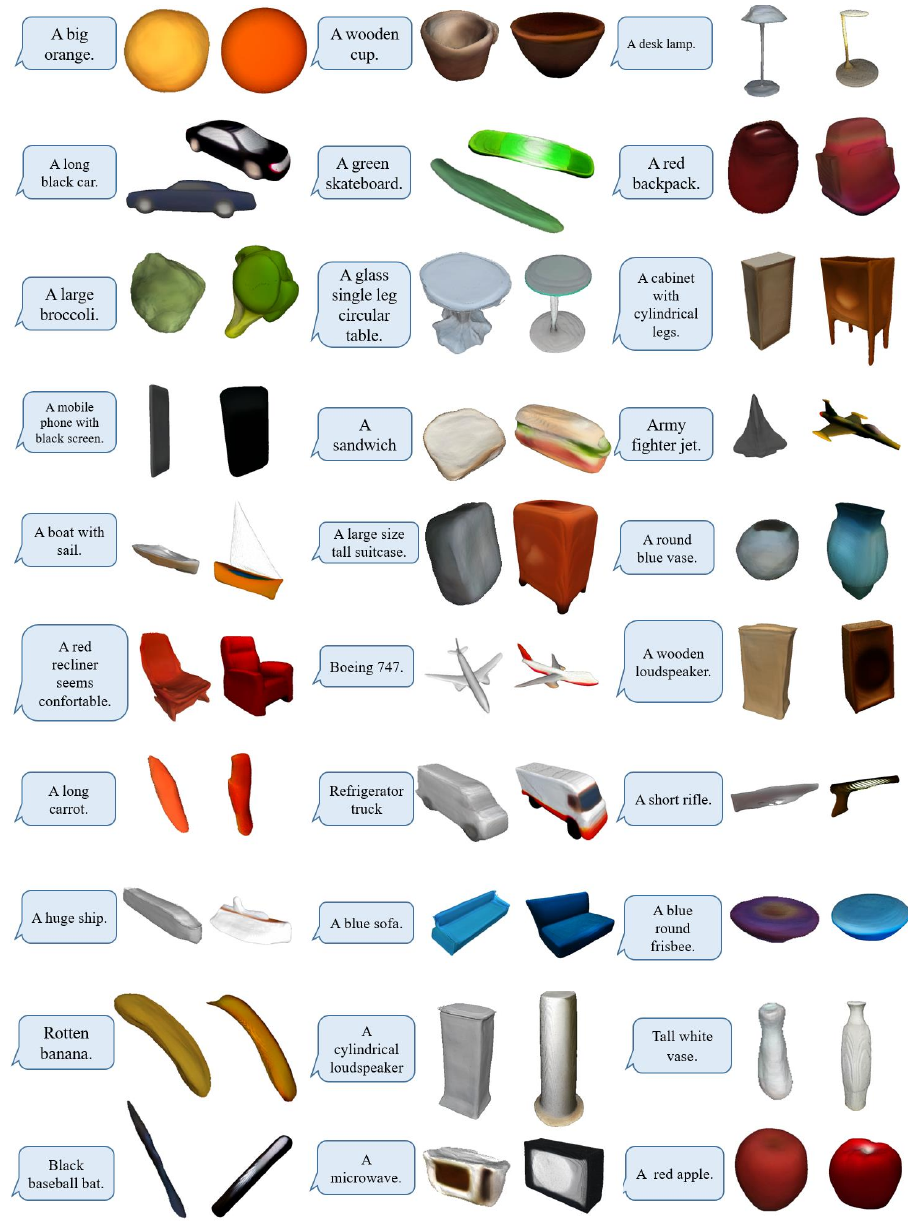}
\vspace*{-1.5mm}
\caption{Generative results of our method. With DreamStone, we can effectively generate shapes of various classes from texts. Left: ISS~\cite{liu2023iss}. Right: DreamStone. Note that DreamStone is built upon ISS~\cite{liu2023iss} and incorporates DP and SDS. 
}
\label{fig:all}
\vspace*{-2.5mm}
\end{figure*}

\vspace*{16pt}
\noindent\textbf{Generation novelty of two-stage feature space alignment.} Our two-stage feature-space alignment has the ability to produce shapes that are novel and not present in the training data. 
Figure~\ref{fig:novelty} shows that given an input text, our model first generates the 3D shape in (a), and then uses it to retrieve the top three closest shapes (b,c,d) in the entire training set based on the cosine similarity between CLIP features $f_\text{I}$ of rendered images. 
The result shows that our generated shapes after two-stage feature space alignment are different from the retrieved shapes, indicating that our two-stage feature space alignment method is able to generate novel shapes even without any stylization process.
It is unsurprising since our two-stage feature space alignment shares the generative space with the
adopted SVR model and has the potential to create all shapes that the adopted SVR model can generate.

\vspace*{10pt}\noindent\textbf{Generation diversity of two-stage alignment.} In Figure~\ref{fig:diversified} and Table~\ref{tab:div}, we compare the diversified generation results of our newly proposed diffusion prior and our previous work ISS~\cite{liu2023iss} both qualitatively and quantitatively. Remember that ISS~\cite{liu2023iss} is also able to generate diversified shapes by randomly perturbating $f_\text{I}$ as initialization and $f_\text{T}$ as the ground truth to derive diversified features. The model can then converge to different shapes for different noise perturbations. 

To evaluate the generative diversity quantitatively, we generate another two shapes per input text for both ISS~\cite{liu2023iss} and DreamStone, then use FID~\cite{heusel2017gans} and FPD~\cite{liu2022towards} for the fidelity and diversity evaluation. Also, we measure the CLIP-Consistency $C=f_T\cdot f_I$ between the rendered images and the input text to evaluate the text-shape consistency. The results in Table~\ref{tab:div} demonstrate that our DreamStone can generate more diversified shapes with better text-shape consistency and quality than ISS~\cite{liu2023iss}. 

Qualitative comparisons are illustrated in Figure~\ref{fig:diversified}. The bounding boxes of different colors indicate the unsatisfactory generative results of our conference version ISS~\cite{liu2023iss}: the red boxes indicate the results that are inconsistent with input texts, the orange boxes indicate the low quality ones, and the blue boxes indicate the too similar and lacking diversity ones; on the contrary, our DreamStone with the diffusion prior mitigates these issues. The results manifest the superior performance of the newly-proposed diffusion prior in terms of generative quality, diversity, and text-shape consistency. 

\begin{table}
\centering
\caption{Quantitative evaluation on diversified generation with diffusion prior. CLIP-C indicates CLIP-Consistency between the rendered images of the generated shapes and the input texts. Note that ``baseline'' indicates ISS~\cite{liu2023iss} and DP (diffusion prior) refers to the model that generates an associated CLIP image feature from the CLIP text feature described in~\cite{ramesh2022hierarchical}.} 
\scalebox{1.2}{
  \begin{tabular}{cccc}
    \toprule
   Method & FID ($\downarrow$) & FPD ($\downarrow$) & CLIP-C  ($\uparrow$) \\
   \midrule
     baseline~\cite{liu2023iss}  & 113.98  & 35.37 & 0.239 \\ 
    baseline~\cite{liu2023iss} + DP & \textbf{108.73}  &  \textbf{34.36} & \textbf{0.248}\\ 
    \bottomrule
  \end{tabular}
  }
  \label{tab:div}
\end{table} 


\vspace*{10pt}\noindent\textbf{Generation fidelity of two-stage feature space alignment.} To evaluate the ability of our two-stage feature space alignment to generate realistic 3D shapes, we train DVR~\cite{niemeyer2020differentiable} on the real-world CO3D dataset, and adopt the learned feature space for text-guided shape generation without using paired data. As depicted in Figure~\ref{fig:co3d}, our model can produce real-world shapes with a high degree of fidelity. To the best of our knowledge, this is the first work to investigate text-guided shape generation on real-world datasets and generate realistic 3D shapes.


\begin{table*}
\centering
\caption{Mean and standard deviation of distances in the feature space mapping process evaluated on our test set. $d$ means cosine distance. Almost all distances are consistently reduced after our stage-2 alignment.  }
\label{tab:mapping}
\scalebox{1.2}{
  \begin{tabular}{cccccc}
    \toprule
     & $d(M(f_\text{I}),M(f_\text{T}))$&$d(M(f_\text{I}), f_\text{S}))$&$d(M(f_\text{T}), f_\text{S}))$&$d(M’(f_\text{T}), f_\text{S}))$&$d(M(f_\text{T}), M’(f_\text{T}))$  \\
    \midrule
mean $\pm$ std&0.58 $\pm$ 0.23&
0.21 $\pm$  0.10&
 0.45  $\pm$  0.20 &
 0.17 $\pm$ 0.08&
 0.32 $\pm$ 0.17\\
    \bottomrule
  \end{tabular}
  }
\end{table*}    

\begin{figure*}
\centering
\includegraphics[width=0.99\textwidth]{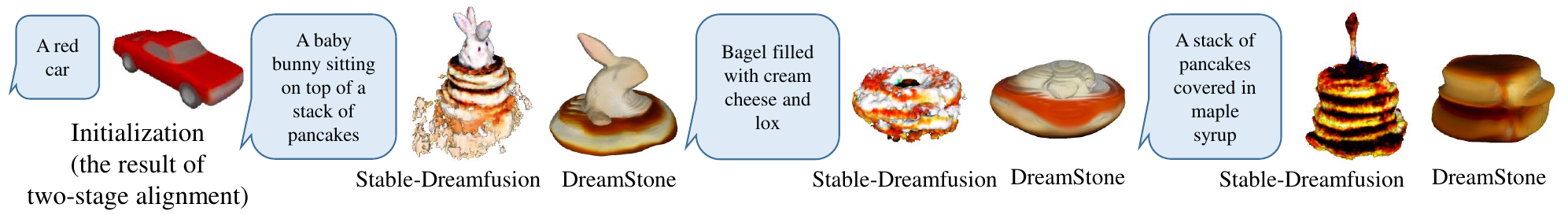}
\vspace*{-2.5mm}
\caption{From a randomly chosen shape (a), some of our out-of-category results (DreamStone) are inferior to DreamFusion (Stable-Dreamfusion). }
\label{fig:limitation}
\vspace*{-3.5mm}
\end{figure*}

\noindent\textbf{Generality and scalability of two-stage feature space alignment on other SVR models.} The generality and scalability of two-stage feature space alignment are evaluated by replacing DVR~\cite{niemeyer2020differentiable} with other SVR models, such as SS3D~\cite{alwala2022pre} and GET3D~\cite{gao2022get3d}. 
It is worth mentioning that SS3D is good at producing 3D shapes in more categories and GET3D is able to generate 3D shapes with higher fidelity.
First, Figure~\ref{fig:ss3d} shows that our approach, built upon SS3D, can generate shapes of more real-world categories, such as birds. 
Notably, the shape generated by our model (left in Figure~\ref{fig:ss3d}(c)) is of better quality than the initial result of the SS3D with an image as input for 3D shape generation from texts. (The right results in Figure~\ref{fig:ss3d}(c) are derived with stage-3 SDS-Guided Refinement, which will be further evaluated in the following Section~\ref{sec:show2}.)
Second, our two-stage alignment is able to fully leverage the generative capabilities of GET3D to produce high-fidelity 3D shapes, as displayed in Figure~\ref{fig:get3d}.
These results demonstrate that our approach is general and compatible with various advanced SVR models for producing shapes of more categories and higher qualities even without SDS-guided refinement.


\subsection{More Analysis of Stage-3 Refinement and Stylization}\label{sec:show2}

Further, we will showcase further text-guided stylization results of Stage-3 refinement and stylization, demonstrating the generality of our generation method on a broad range of categories.

\vspace*{10pt}
\noindent\textbf{Generation beyond the capability of the SVR model.} The text-guided stylization module enables our model to create 3D shapes beyond the pre-trained SVR model. As shown in Figure~\ref{fig:style_figure1}, Figure~\ref{fig:overview_style}, Figure~\ref{fig:style}, 
and Figure~\ref{fig:sds_style}, novel structures and textures matching text descriptions can be created. In Figure~\ref{fig:style}, we present stylization results from our three stylization approaches using the same text prompts. 
As shown in Figure~\ref{fig:style} (a), the CLIP-guided texture stylization can hallucinate both realistic (``mahogany chair'') and fantasy (``glacier chair'') vivid textures on the chair. Also, it can create higher-fidelity textures than the other stylization techniques and can better handle abstract text descriptions; see Figure~\ref{fig:style_compare} (a) ``sunset''. 
Further, in Figure~\ref{fig:style} (b), our shape-and-texture stylization successfully creates novel textures and imaginary shapes not present in the training dataset. 
Also, as shown in Figures~\ref{fig:style_figure1},~\ref{fig:style} (c), and~\ref{fig:sds_style}, our DreamStone is capable of generating aesthetically pleasing stylized shapes with intricate details and textures, such as the ``rabbit lamp'' and ``banana chair''. These results showcase the ability of our model to generate visually appealing and complex shapes from text descriptions.


\vspace*{10pt}\noindent\textbf{More generative results.} In addition, we showcase a diverse range of 3D shapes that have been effectively generated from texts using our approach in Figure~\ref{fig:all}.

\subsection{Analysis of Feature Space Mapping}\label{sec:mapping}

To better understand how our two-stage feature-space alignment works, we further study the average feature distances at different stages for all samples in our test set as shown in Table~\ref{tab:mapping}. Please also refer to Figure~\ref{fig:motivation} (c) for the visualized results. 

In the stage-1 alignment, we train the mapper $M$ to map the CLIP image feature $f_\text{I}$ to $M(f_\text{I})$ that is close to the target shape $f_\text{S}$ with latent-space regression. Based on the fact that the CLIP model is able to map $f_\text{I}$ and $f_\text{T}$ to a shared embedding space, it is a natural assumption that the mapper $M$ is also able to map $f_\text{T}$ close to the target shape space. Yet, we found that there is a large gap between $M(f_\text{T})$ and $f_\text{S}$ even with the stage-1 alignment. Specifically, the average distance of all samples between $M(f_\text{T})$ and $M(f_\text{I})$ is $0.58 \pm 0.23$, indicating a substantial gap between the CLIP image and text features. Also, the measured average distance between $M(f_\text{I})$ and $f_\text{S}$ is $0.21 \pm 0.10$, while the distance of mapped text and shape is $d(M(f_\text{T}), f_\text{S}) = 0.45 \pm 0.20$, indicating a large room for further improvement. Importantly, the above motivates us to adopt an additional stage-2 alignment. It should be noted that since there is no ground truth 3D shape in our task, we manually select a shape from the ShapeNet dataset that matches well with the input text as the ground-truth one.

During the stage-2 alignment, the mapper $M$ is fine-tuned to be $M'$ for each input text to further narrow the gap between $M'(f_\text{T})$ and $f_\text{S}$ to be $0.17 \pm 0.08$, which is much smaller than $0.45 \pm 0.20$, i.e., $d(M(f_\text{T}), f_\text{S})$ after the stage-1 alignment. This analysis manifests that the stage-2 alignment can effectively reduce the gap between features of the mapped text and reference shape.

\section{Limitations}
\label{sec:limitation}
DreamStone trades off between the generative fidelity of 3D shapes within the image dataset and the generation capability for categories outside the image dataset. Though DreamStone can generate shapes outside the dataset with better surface quality (as shown in Figure~\ref{fig:out-category}), its out-of-category generative capability may not always outperform DreamFusion~\cite{poole2022dreamfusion}, as shown in Figure~\ref{fig:limitation}. We empirically found that it can be helpful to choose an initialization shape of a similar topology as the desired shape for the SDS procedure. Yet, there is still a lack of guidance on how to choose a suitable initialization shape for an out-of-category generation.

\section{Conclusion}
In this work, we introduce a novel approach for text-guided 3D shape generation that leverages the image modality as a stepping stone. Our approach eliminates the need for paired text and shape data by using joint text-image features from CLIP and shape priors from a pre-trained single-view reconstruction model. Technically, we have the following contributions.
First, our two-stage feature-space alignment reduces the gap between text, image, and shape modalities. Second, the text-guided refinement and stylization techniques effectively and efficiently equip the generated 3D shapes with rich details and diverse styles. Third, our proposed approach is compatible with different single-view reconstruction methods and can be developed to produce shapes in a wide variety of categories and with higher fidelity. 
Experimental results on ShapeNet, CO3D, and additional categories demonstrate that our approach outperforms SOTA approaches and various baselines.

%

\section*{ACKNOWLEDGEMENTS}
The work has been supported in part by  the Research Grants Council of the Hong Kong Special Administrative Region (Project no. CUHK 14206320), General Research Fund of Hong Kong (No. 17202422), Hong Kong Research Grant Council - Early
Career Scheme (Grant No. 27209621), General Research Fund (Grant No. 17202422) and National Natural Science Foundation of China (No. 62202151). We would also like to thank Mr. Jingyu Hu from The Chinese University of Hong Kong and Dr. Karsten Kreis from NVIDIA's Toronto AI Lab for insightful discussions and contributions to the ideas presented in this work.



%


\bibliographystyle{ieee}
\bibliography{egbib}

\begin{IEEEbiography}
[{\includegraphics[width=1in,height=1.25in,clip,keepaspectratio]{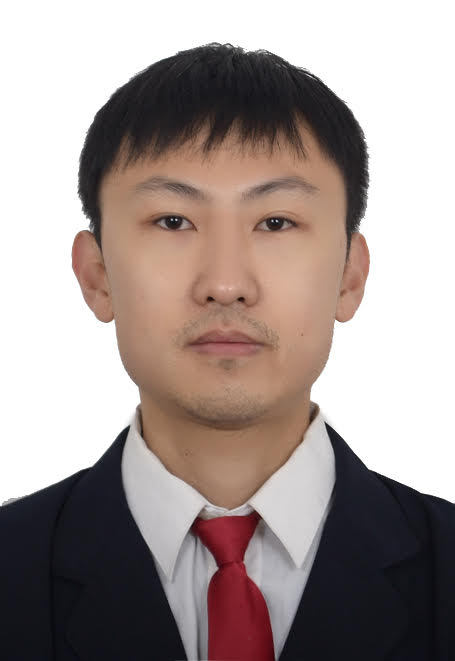}}]
{Zhengzhe Liu} is currently a Ph.D. candidate at The Chinese University of Hong Kong. He received his B.Eng degree in Information Engineering at Shanghai Jiao Tong University, and the M.Phil. degree in Computer Science and Engineering from The Chinese University of Hong Kong. His research interests include AIGC, 3D shape generation, and 3D scene understanding. 
\end{IEEEbiography}

\begin{IEEEbiography}
[{\includegraphics[width=1in,height=1.25in,clip,keepaspectratio]{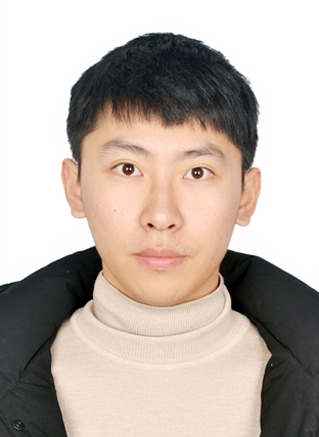}}]
{Peng Dai} received the B.Eng. and M.Eng. degrees from the University of Electronic Science and Technology of China, in 2017 and 2020, respectively. He is currently a Ph.D. candidate at the University of Hong Kong.  His research interests lie at computer vision, computer graphics, and neural rendering.
\end{IEEEbiography}

\begin{IEEEbiography}
[{\includegraphics[width=1in,height=1.25in,clip,keepaspectratio]{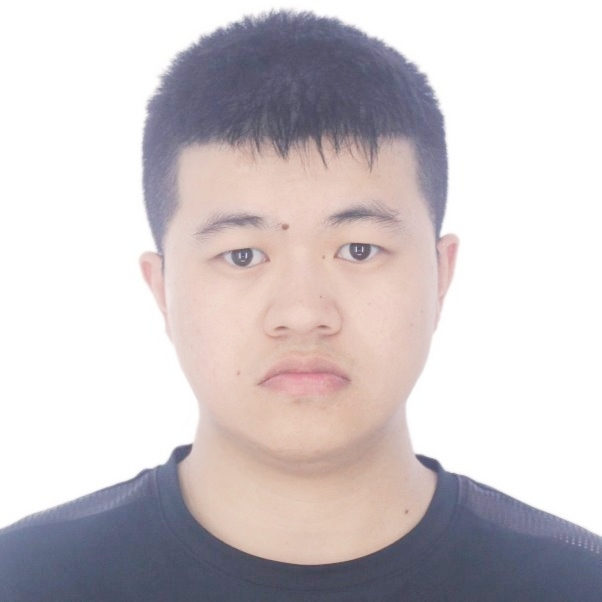}}]
{Ruihui Li} is currently an associate professor at Hunan University. Before that, he was a post-doctoral fellow at the Chinese University of Hong Kong. He received his Ph.D. degree in the Department of Computer Science and Engineering from the Chinese University of Hong Kong.  His research interests include deep geometry learning, generative modeling, 3D vision, and computer graphics.
\end{IEEEbiography}

\begin{IEEEbiography}[{\includegraphics[width=1in,height=1.25in,clip,keepaspectratio]{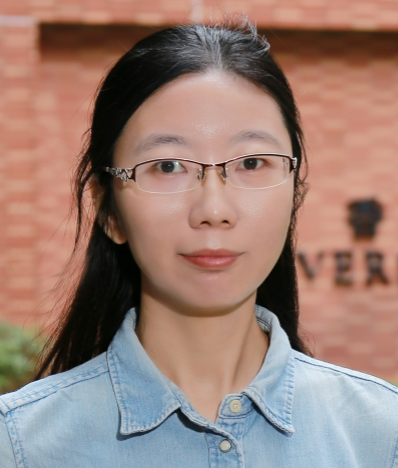}}]{Xiaojuan Qi} is currently an assistant professor at the University of Hong Kong and a member of Deep Vision Lab. Before that, she
received her B.Eng degree in Electronic Science and Technology at Shanghai Jiao Tong University (SJTU) in 2014, and the PhD degree in Computer Science and Engineering from the Chinese University of Hong Kong in 2018. Her research lies in the broad areas of Computer Vision, Deep Learning, and Artificial Intelligence. 
\end{IEEEbiography}

\begin{IEEEbiography}[{\includegraphics[width=1in,height=1.25in,clip,keepaspectratio]{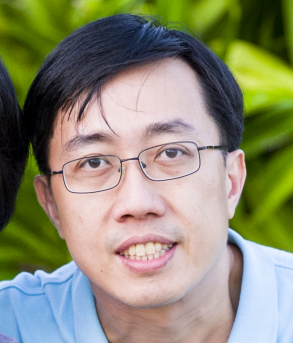}}]{Chi-Wing Fu} is currently a full professor at the Chinese University of Hong Kong.  He is now serving as the Associate Editor-in-Chief (regular submissions) of IEEE Computer Graphics and Applications.  He served as the program co-chair of SIGGRAPH ASIA 2016 technical brief and poster, associate editor of Computer Graphics Forum and IEEE Computer Graphics and Applications, and program committee member in various conferences such as SIGGRAPH, IEEE Visualization, and IEEE VR.  His research interests include computer graphics, 3D vision, user interaction, and visualization.
\end{IEEEbiography}




\end{document}